\documentclass[journal]{IEEEtran}
\usepackage[dvipsnames]{xcolor}
\usepackage{amsmath,amstext,epsfig,diagbox,multirow,multicol,graphicx,subfloat}
\usepackage{pdfpages}
\usepackage{amsmath,amssymb} 
\usepackage{cite}
\usepackage{tikz}
\usepackage{subcaption}
\usetikzlibrary{fit,positioning}
\tikzset{isometricXYZ/.style={x={(-0.866cm,-0.5cm)}, y={(0.866cm,-0.5cm)}, z={(0cm,1cm)}}}
\graphicspath{{images/}}
\ifCLASSINFOpdf{}
\else
\fi

\begin{document}
\title{Generative Model with Coordinate Metric Learning for Object Recognition Based on 3D Models}

\author{\IEEEauthorblockN{Yida Wang\IEEEauthorrefmark{1} and
		Weihong Deng\IEEEauthorrefmark{1},~\IEEEmembership{Member,~IEEE}
	} \\
	\IEEEauthorblockA{\IEEEauthorrefmark{1}School of Information and Communication Engineering,
	Beijing University of Posts and Telecommunications, Beijing, BJ 100876 China}
	\thanks{
Corresponding author: W Deng (email: whdeng@bupt.edu.cn).}}

\markboth{}%
{Wang \MakeLowercase{\textit{et al.}}: Bare Demo of IEEEtran.cls for IEEE Journals}
\maketitle

\begin{abstract}
Collecting data for deep learning is so tedious which makes it hard to establish a perfect database.
In this paper, we propose a generative model trained with synthetic images rendered from 3D models which can reduce the burden on collecting real training data and make the background conditions more sundry.
Our architecture is composed of two sub-networks: semantic foreground object reconstruction network based on Bayesian inference and classification network based on multi-triplet cost training for avoiding over-fitting on monotone synthetic object surface and utilizing accurate informations of synthetic images like object poses and lightning conditions which are helpful for recognizing regular photos.
Firstly, our generative model with metric learning utilizes additional foreground object channels generated from semantic foreground object reconstruction sub-network for recognizing the original input images.
Multi-triplet cost function based on poses is used for metric learning which makes it possible training an effective categorical classifier purely based on synthetic data.
Secondly, we design a coordinate training strategy with the help of adaptive noises applied on inputs of both of the concatenated sub-networks to make them benefit from each other and avoid inharmonious parameter tuning due to different convergence speed of two sub-networks.
Our architecture achieves the state of the art accuracy of 50.5\% on ShapeNet database with data migration obstacle from synthetic images to real photos.
This pipeline makes it applicable to do recognition on real images only based on 3D models.
\end{abstract}


%
\IEEEpeerreviewmaketitle{}

\section{Introduction\label{sec:intro}}

\begin{figure}[ht]
	\captionsetup[subfigure]{justification=centering}
	\centering
	\begin{subfigure}[t]{1\linewidth}
		\centerline{\includegraphics[width=1\linewidth]{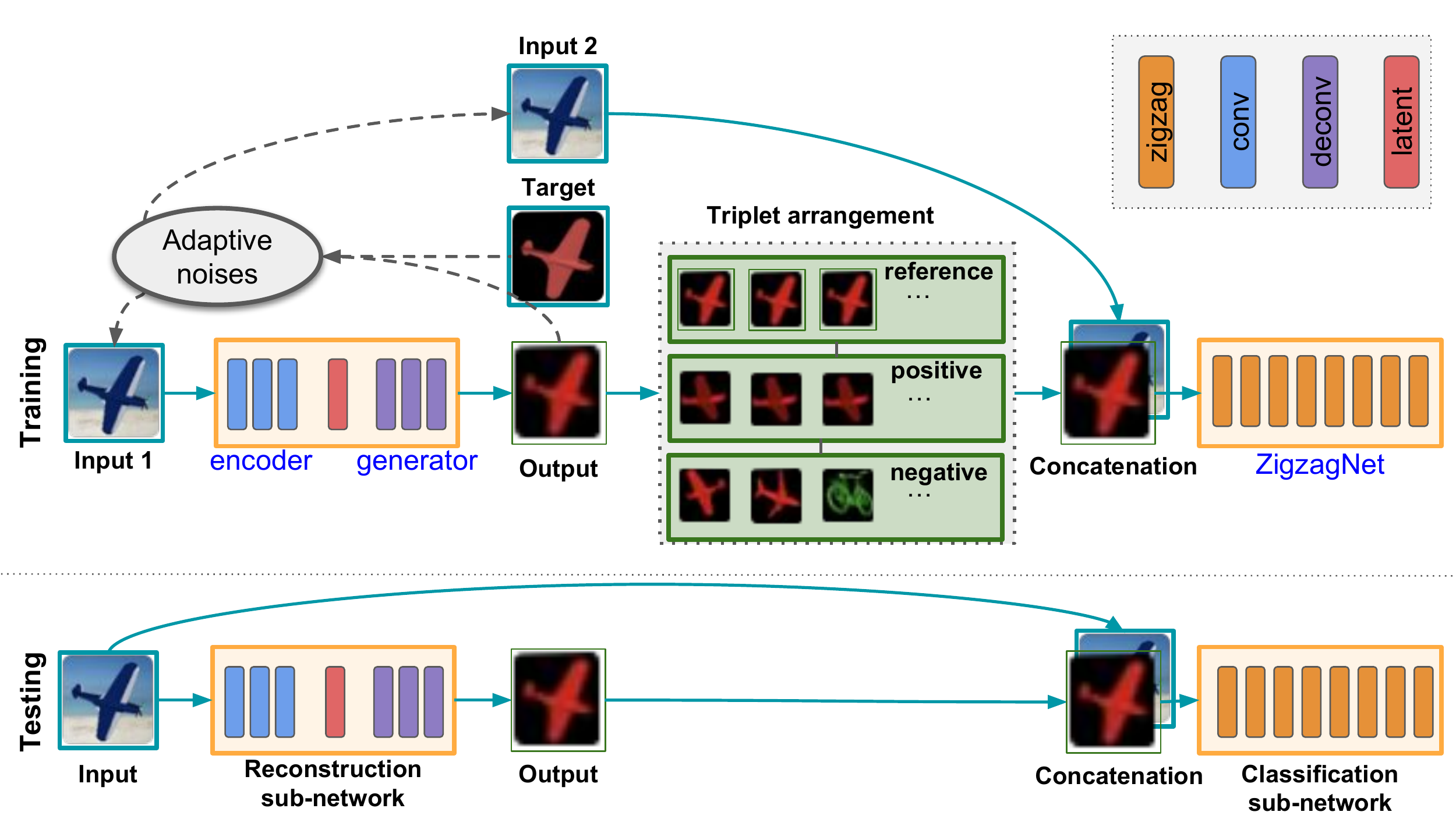}}
		\caption{Flowchart of generative model with coordinate metric learning using 3D models. \textbf {Top:} training stage. \textbf{Bottom:} testing stage.\label{fig:pipeline_tip}}
	\end{subfigure}
	\begin{subfigure}[t]{0.48\linewidth}
		\centerline{\includegraphics[height=1.6cm]{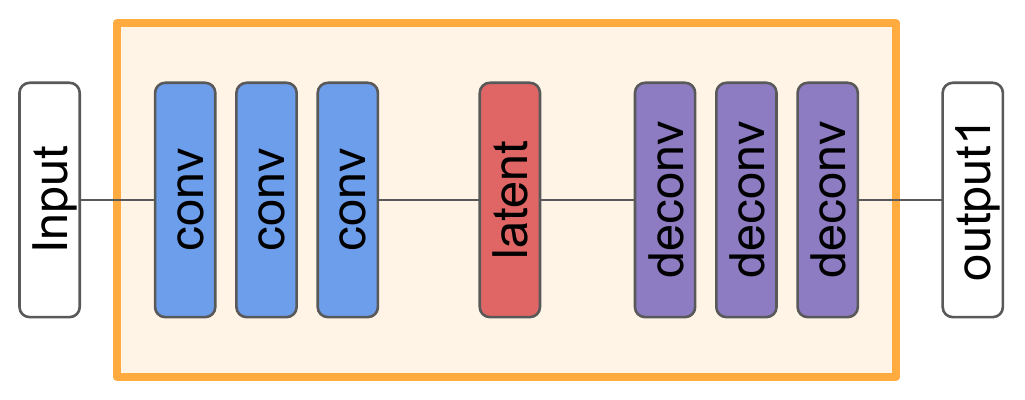}}
		\caption{Reconstruction network\label{fig:network_rec}}
	\end{subfigure}
	\begin{subfigure}[t]{0.48\linewidth}
		\centerline{\includegraphics[height=1.6cm]{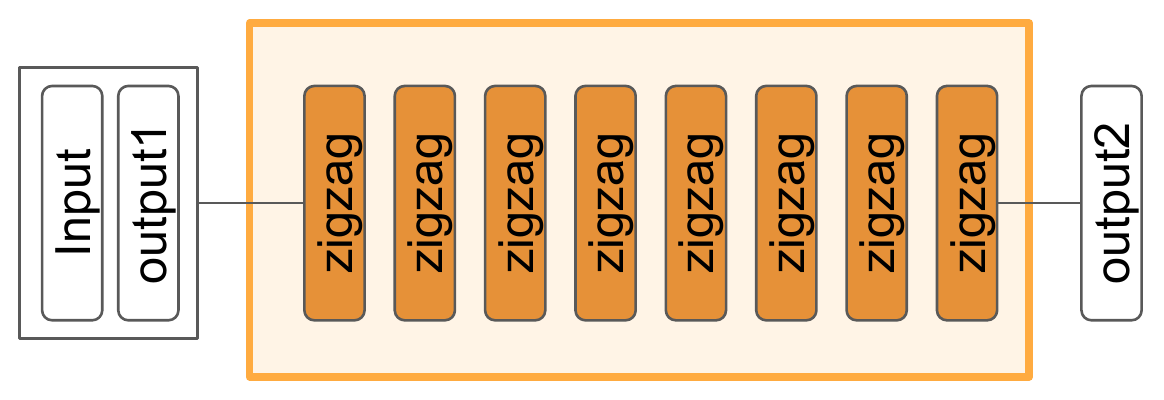}}
		\caption{Classification network\label{fig:network_cls}}
	\end{subfigure}
	\caption{Generative model with coordinate metric learning.\label{gm_cml}}
\end{figure}

\IEEEPARstart{D}{eep} architectures based on convolutional operations such as AlexNet~\cite{Krizhevsky2012}, Inception concept~\cite{SzegedyLJSRAEVR14} and ResidualNet~\cite{he2015deep} work well on image classification trained on large scale real datasets such as ImageNet~\cite{Olga2014} and PASCAL~\cite{everingham2015the}.
Meanwhile, synthetic data could also be served as training data for specific tasks such as pose estimation~\cite{wohlhart15, corrSuQLG15}.
Triplet cost function~\cite{wohlhart15} used in metric learning utilizes these accurate labels well by making arrangements for triplet set according to corresponding labels of synthetic data, so manifold learning is more applicable based on synthetic data than realistic data.
Pose labels are also useful for exploiting geometric information, recent works on object rotation and deformation combining CNN with auto-encoder~\cite{Yim2015Rotating, NIPS2015_5639} solve tasks such as face rotation and intrinsic transformations for objects based on additional pose information.
Other supervision information such as remote codes~\cite{Yim2015Rotating} gives a help on image transformation in regard to different purposes which modify auto-encoder to flexible reconstructor.

Though typical methods are effective in multi-task learning such as pose estimation~\cite{wohlhart15} and categorical classification~\cite{PepikBRS15} with additional information recorded by special equipments,
performance of testing on real photos with severe data migration problem compared to synthetic training data is still not satisfied.
Works on foreground object segmentation~\cite{wang_icip2016} try to solve data migration problem from synthetic training data and real testing photos concatenating segmentation channel to original photo as input for classification architecture,
but those concatenated architecture still have over fitting problem on original RGB channels without fully utilization on robust object segmentation result.

In this paper, we propose a deep architecture targeted on training with synthetic data and recognizing on real photos especially for classification.
Such graph concatenated with two sub-networks which are reconstruction network based on Bayesian inference and classification network based on ZigzagNet.
Our reconstruction architecture trained from two types of synthetic data rendered from models of ShapeNet~\cite{shapenet2015} database reconstructs foreground objects from background with surface colors representing categories.
Inspired by positive impact of depth images for pose estimation~\cite{wohlhart15} and face rotation~\cite{Yim2015Rotating}, three additional channels provided by object reconstruction sub-network are concatenated to original RGB channels altogether as a six-channel data feeding for classification sub-network.
Pixel to pixel prediction is possible based on encoding and decoding architecture such as Fully Convolutional Network (FCN)~\cite{Long2015Fully, Wang2015Visual}, Deconvolutional Network (DeconvNet)~\cite{Noh2015Learning} and Variational Auto-encoder (VAE)~\cite{Kingma2013, Blei2016Variational}.
Inspired by good performance introduced by statistical latent variables in generative model~\cite{Blei2016Variational}, we modify conditional VAE~\cite{NIPS2015_5775, Kingma2014Semi} to make the reconstruction sub-network able to generate semantic foreground objects robustly without feeding in additional supervision information for testing.
Textured models from ShapeNet~\cite{shapenet2015} database are used for avoiding over-fitting on monotone surface of texture-less models such as models in PASCAL 3D+~\cite{xiang_wacv14}.
Our classification sub-network is optimized with a joint cost function composed of categorical softmax loss and pose assisted triplet loss. Such optimization tries to avoid over fitting on the unrealistic relationship between synthetic objects and background by means of learning discriminant geometric features according to different camera poses.
Gradients from our joint cost function for object classification is back-propagated to object reconstruction sub-network which is the head of concatenated architecture as a supervision information like remote codes~\cite{Yim2015Rotating} to help reconstruct object in different categories accordingly.
Our conjugate network aiming at reconstructing foreground objects and doing classification on categories in the same time is trained efficiently with adaptive noises for both inputs of two sub-networks.
With the help of coupled noises calculated from variances of reconstructed output and ground truth mask applied on both inputs of two sub-networks,
problem of gradient vanishing from classification network back to the output channels of reconstruction network is avoided.
This means that the reconstructed object masks are more useful for training classification network as additional information.

Experiments on categorical prediction task for objects in real images based on photos provided in PASCAL 3D+~\cite{xiang_wacv14} show that our concatenated architecture trained merely on synthetic data solves severe data migration problem from synthetic training to realistic testing data well which achieves the state of the art accuracy of 50.5\% on ShapeNet database without using Nearest Neighbor classifier in LDO~\cite{wohlhart15}.
Utilization on pose information with our joint cost function makes the classification accuracy much closer to the ones trained on real photos than some recent works like LDO~\cite{wohlhart15} and SR~\cite{wang_icip2016}.

In the following chapters, we describe our method for training deep architecture merely based on 3D models aiming for recognizing objects in real photos.
Our theory could be roughly divided into those parts:
	\begin{itemize}
		\item General rendering strategy from 3D models to 2D images is described in Section~\ref{sec:data_prep}. Additional information prepared for metric learning such as poses are also recorded while generating images.
		\item Section~\ref{sec:recon_generative} describes our generative model targeted on pixel-wise reconstruction which provides additional information including segmentation masks with semantic colors and depth information.
		\item In Section~\ref{sec:classification}, we describe the second part of the conjugate architecture. Classification sub-network is modified into compact structure based on ZigzagNet in Section~\ref{sec:classification_compact_arch} to avoid over fitting.
		\item Furthermore, parametric model of classification sub-network is also optimized with metric learning in Section~\ref{sec:classification_metric} supervised by category labels and pose information which also make it possible back-propagating effective gradients to reconstruction sub-network.
		\item Finally, paired adaptive noises contribute to a coordinate training process for both sub-networks by means of making adaptive corruptions on both of the RGB inputs channels which is described in Section~\ref{sec:coordinate}.
	\end{itemize}

\section{Related Works\label{sec:related_works}}

Some methods related to training on synthetic data generated with rendering techniques~\cite{Aubry14, PengSAS14} are useful for tasks in specific conditions such as pose estimation~\cite{wohlhart15, corrSuQLG15} in indoor conditions because camera positions are used for image rendering which could be recorded accurately.
Although synthetic data could hardly be as realistic as real photos, some methods still utilize more information to overcome data migration problem for classification.
For example, SR~\cite{wang_icip2016} utilize two types of synthetic data for training a concatenated architecture for final task of classification.
One core task in this paper is utilizing 3D models with rich information for training deep parametric model which could be applied on real world visual recognition tasks.
We train deep neural network directly from images rendered from 3D models and test on real photos, especially for classification.

Metric learning is helpful to learn and exploit implicit relationships among data samples.
Descriptor-based metric learning methods such as Triplet training~\cite{wohlhart15} and Siamese training~\cite{lu2014neighborhood} utilize additional informations for person re-identification, 3D object pose estimation and kinship verification by learning specific manifold in the embedded descriptor space.
Metric learning on embeddings such as~\cite{oh2016deep} uses pair-wise distances which is helpful for retrieval tasks.
Joint Siamese and Triplet training is also applied on local descriptor optimization for patch matching~\cite{kumar2016learning}.
Here we utilize a multi-triplet loss for optimizing a concatenated architecture to make it applicable for training on synthetic data and testing on real data.

\section{Training Data Generation\label{sec:data_prep}}

The first mission is representing 3D information in 2D space for training data generation based on 3D models.
We construct a new database which consists of two types of synthetic images rendered from 3D models of ShapeNet~\cite{shapenet2015} database which are the one including textured objects with background images crawled from Flickr and another one including textureless objects without any background.
The second one will be used as semantic object masks in training stage, so we render the objects with monotone color in regard to specific category.
Both types of synthetic images are rendered with the same camera parameters accordingly.

\subsection{Semantic Rendering\label{sec:data_prep_render}}

Inspired by the fact that depth images can provide complementary information for recognition on real images~\cite{wohlhart15} by distinguishing foreground target from background, we embed an semantic foreground object reconstruction sub-network in the whole generative model based on Bayesian coding.
The output of the reconstruction network is concatenated to the original images as a six channel data forwarded to the classification network.
These concatenated channels reconstructed from realistic input images with background are expected to be similar to segmented object image with unique colors according to category.

As shown in Fig.~\ref{fig:pipeline_tip}, our goal is training a robust semantic rendering sub-network which provides additional information for classification sub-network with the aim similar to the one of the self-restraint foreground object reconstruction network~\cite{wang_icip2016}.
The reconstructed semantic masks with depth information is concatenated to the original RGB channels together as a 6 channel input to the next classification network.
Such concatenation also makes the statistical gradient of the classifier able to be propagated back to the reconstructed network.
Reconstructed channels for foreground objects are expected to contain segmentation information on contours, depth information and semantic color information which means that they do not only reflect informative details about the foreground objects, but also help to train the classification network with visual categorical information.

\begin{figure}[ht]
	\captionsetup[subfigure]{justification=centering}
	\centering
	\begin{subfigure}[t]{0.46\linewidth}
		\begin{tikzpicture} [scale=1.4, isometricXYZ, line join=round,
			opacity=.75, text opacity=1.0,%
			>=latex,
			inner sep=0pt,%
			outer sep=2pt,%
			]
			\def\h{5}

			\newcommand{\quadrant}[2]{\foreach \f in {85,75,65,55,45,35,25,15,5}
				\foreach \t in {#1}
				\draw [dotted, fill=#2]
				({sin(\f - \h)*cos(\t - \h)}, {sin(\f - \h)*sin(\t - \h)}, {cos(\f - \h)})
				-- ({sin(\f - \h)*cos(\t + \h)}, {sin(\f - \h)*sin(\t + \h)}, {cos(\f - \h)})
				-- ({sin(\f + \h)*cos(\t + \h)}, {sin(\f + \h)*sin(\t + \h)}, {cos(\f + \h)})
				-- ({sin(\f + \h)*cos(\t - \h)}, {sin(\f + \h)*sin(\t - \h)}, {cos(\f + \h)})
				-- cycle;
			}

			\newcommand{\arrowarc}[6]{\draw[green,domain=0:320,smooth,variable=\x,-] plot
				({0.07 * (cos(#2)*cos(#3) * cos(\x) + (cos(#3)*sin(#1)*sin(#2) - cos(#1)*sin(#3)) * sin(\x)) + #4},
				{0.07 * (cos(#2)*sin(#3) * cos(\x) + (cos(#1)*cos(#3) + sin(#1)*sin(#2)*sin(#3)) * sin(\x)) + #5},
				{0.07 * (-sin(#2) * cos(\x) + cos(#2)*sin(#1)* sin(\x)) + #6});
			}

			\quadrant{130,150,170,190,210,230,250,270,290,310}{black!2}
			\quadrant{-50,-30,-10,10,30,50,70,90,110,130}{black!2}

			\foreach \f in {50,60,70,80,90}
			\foreach \t in {-40,-20,0,20,40,60,80,100,120,140,160,180}
			{\def\l{1.15}
				\draw [red, ->, thin]
				({\l*sin(\f)*cos(\t)},{\l*sin(\f)*sin(\t)},{\l*cos(\f)})
				-- ({sin(\f)*cos(\t)},{sin(\f)*sin(\t)},{cos(\f)});

				\def\l{1.12}
				\arrowarc{(\f)}{0}{(\t + 90)}{\l*sin(\f)*cos(\t)}{\l*sin(\f)*sin(\t)}{\l*cos(\f)}
			};
		\end{tikzpicture}
		\caption{Camera ranges\label{fig:camera_ranges}}
	\end{subfigure}
	\begin{subfigure}[t]{0.47\linewidth}
		\begin{tikzpicture}[scale=0.5]
			\tikzstyle{every node}=[circle,fill=green!15,inner sep=0pt]
			\foreach \y[count=\a] in {10,9,4}
			{\pgfmathtruncatemacro{\kn}{120*\a-90}
			\node at (\kn:3) (b\a) {\small \y};}
			\foreach \y[count=\a] in {8,7,2}
			{\pgfmathtruncatemacro{\kn}{120*\a-90}
			\node at (\kn:2.2) (d\a) {\small \y};}
			\foreach \y[count=\a] in {1,5,6}
			{\pgfmathtruncatemacro{\jn}{120*\a-30}
			\node at (\jn:1.5) (a\a) {\small \y};}
			\foreach \y[count=\a] in {3,11,12}
			{\pgfmathtruncatemacro{\jn}{120*\a-30}
			\node at (\jn:3) (c\a) {\small \y};}
			\draw[dashed] (a1)--(a2)--(a3)--(a1);
			\draw[thick] (d1)--(d2)--(d3)--(d1);
			\foreach \a in {1,2,3}
			{\draw[dashed] (a\a)--(c\a);
			\draw[thick] (d\a)--(b\a);}
			\draw[thick] (c1)--(b1)--(c3)--(b3)--(c2)--(b2)--(c1);
			\draw[thick] (c1)--(d1)--(c3)--(d3)--(c2)--(d2)--(c1);
			\draw[dashed] (b1)--(a1)--(b2)--(a2)--(b3)--(a3)--(b1);
		\end{tikzpicture}
		\caption{Initial vertexes on icosahedron\label{fig:icosahedron}}
	\end{subfigure}
	\caption{Synthetic camera positions on the vertexes of a sphere.\label{fig:cameras_tip}}
\end{figure}
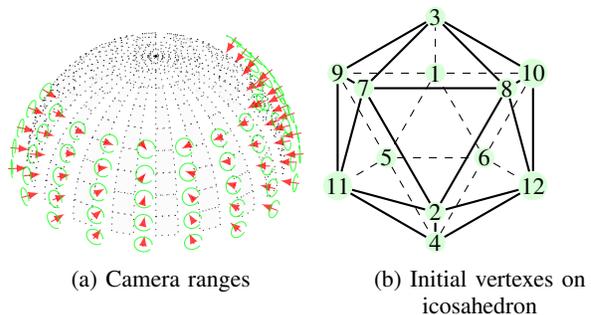

2D synthetic training images are rendered from 3D models in database with texture files like ShapeNet~\cite{shapenet2015}.
Foreground objects are shot from vertexes on a sphere net combined with triangles recursively generated from larger triangles which are initiated from a regular icosahedron~\cite{BMVC.22.10} shown in Fig.~\ref{fig:icosahedron}.
we cut off the upper and lower parts of the whole sphere along with two meridians vertical to $z$ axis cause range of view positions are always limited in real world.
For the coordinate $ (V_x, V_y, V_z) $ of camera positions shown in Fig.~\ref{fig:camera_ranges}, $ V_z \in [-0.1 \min(V_x), 0.6 \max(V_x)] $.
Focal points are set randomly within a range around the center of each object to simulate photos taken as close shots.
As we want to utilize realistic data to train a semantic foreground objects reconstruction network afterwards,
background images are crawled from \textit{Flickr} where target objects are not included.

For fine-tuning with more data, we also render another collection of close shot images with shifted focal points which are moved from the center of object to the head of objects.
Shifting distance from the original focal points in the center could be represent as $ (F_x', F_y', F_z') = (F_x, F_y, F_z) + 0.2((C_x, C_y, C_z)-P_{axis})$ where $ F $ represents the focal points used for the camera shots, $ C $ represents the camera position on the sphere and $ P_{axis} $ represents the intersection point of the axis of front view and the sphere net of camera positions.

\begin{figure}[ht]
	\captionsetup[subfigure]{justification=centering}
	\centering
	\centering
	\centerline{\includegraphics[width=1\linewidth]{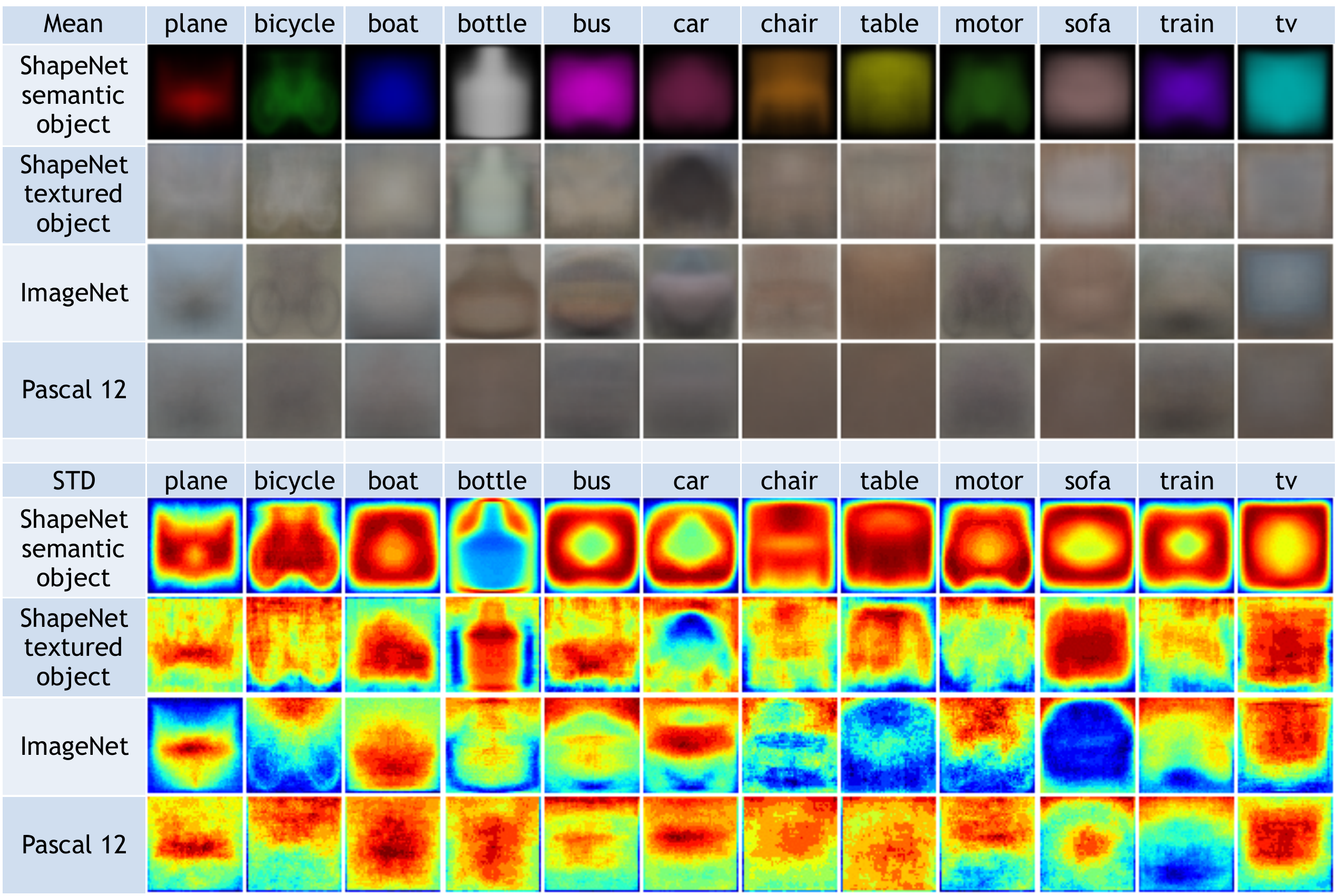}}
	\caption{Mean value and normalized heat map for standard deviation of images including 2 types of synthetic images (upper 2 rows) and 2 types of real photos (lower 2 rows) in large scale database.\label{fig:imganalysis}}
\end{figure}

Mean and normalized standard deviation images from ShapeNet and ImageNet in Fig.~\ref{fig:imganalysis} show that gap between synthetic ones and real ones is still too large to train a recognition model directly from synthetic data.
A challenging task is overcoming data migration problem between synthetic data and real photos because we want to get rid of the over-fitting problem introduced by our realistic data.

\section{Semantic Reconstruction\label{sec:recon}}

Our semantic reconstruction removes background environments from foreground objects and renders them with specific color according to specific category for further classification, so images with textured foreground objects will have the same monotone surface color if they are coming from the same category.
Reconstruction sub-network is trained from paired synthetic images which are realistic images with background pictures and semantic segmentation masks of foreground objects.
We apply two methods to make the reconstruction network capable on semantic rendering for foreground objects.
Firstly, we design a generative model using Bayesian inference to make the generator more capable on perceiving global information on objects compared to traditional pixel-wise coding methods like FCN~\cite{Long2015Fully} and DeconvNet~\cite{Noh2015Learning}.
Other supervision assistants for reconstruction like remote code described in face rotating~\cite{Yim2015Rotating} for multi-task learning according to the identities of objects are discarded while our gradients of joint loss acting as supervision in classification sub-network is back-propagated into the reconstruction network.
Then adaptive noises are applied automatically on input of the reconstruction sub-network to fully utilize the back-propagated gradients from classification sub-network sequentially concatenated with the generated output of reconstruction sub-network,
so our whole architecture is called generative model with coordinate metric learning.
Both of the conjugate sub-networks are coupled with each others by utilizing the adaptive noises.

\subsection{Pixel-wise Generative Model\label{sec:recon_generative}}

Let us suppose $X$ to represent the input samples. 
Our probabilistic variational framework is composed of two parts: an encoder $q(X)$, that compresses each input $X$ into latent variables $z$, and a generator or decoder $p(z)$ that maps $z$ into the output $Y$.
Generally speaking, our model could be regarded as a variational coder (VC) specifically designed for image related tasks.

We represent the prediction task $p(q(X))$ as a stochastic process modeled by the two probability distributions $P(Y|z)$ and $P(z)$\footnote{We use capital letters for representing probability density function and lower case letters for normal functions which is deterministic from input to output.}.

Inspired by VAE, a Gaussian distribution $P(z) \sim N(0,I)$ with the identity matrix $I$ is adopted for modeling the latent variables $z$, this meaning that the likelihood $P(Y|z)$ also follows a Gaussian distribution: $Y \sim N(p(z),\sigma^2*I)$ with covariance defined by the scalar $\sigma$ having the same dimension as the number of latent variables.

One of the main ideas behind our approach is to optimize both the encoder $p(X)$ and the generator $p(z)$.
The negative log probability of the output $Y$ is proportional to the squared Euclidean distance between the output $p(z)$ and the expected target $Y$. Therefore, the main task is how to optimize $p(X)$.
Since in theory the latent variables $z$ should be able to reproduce any output $Y$,
the target of the optimization procedure for the encoder function $p(X)$ is set to be the matching posterior $P(z|Y)$, which means that the latent variables are more likely to produce expected outputs.

Firstly, we need to apply Bayesian rule
\begin{align}
	\label{equ:bayesian1}
	P(z|Y) = \frac{P(Y|z)P(z)}{P(Y)} = \frac{P(Y|z)P(z)}{\int P(Y|z) P(z) dz}
\end{align}
to calculate the true posterior $P(z|Y)$ from training samples.
The problem is that the range of $z$ is not limited at all, so the likelihood $P(Y|z)$ is almost zero for most $z$. Thus, we can not calculate this likelihood directly from the samples.
To solve this problem, we set some constraints on the latent variables $z$ to make them more likely to generate $Y$.
Meanwhile, since the optimization of the encoder $p(X)$ suffers a similar limitation induced by the limited output range,
we redefine the encoder function as $q(X)$, an alternative for $p(X)$, such that the associated probability density function $Q(z|X)$ is expected to match the posterior $P(z|Y)$.

Under these conditions, the expectation of the likelihood $E_{z\sim Q}P(Y|z)$ conditioned on the latent variables should be close to the true probability $P(Y)$. Kullback-Leibler (KL) divergence between $P(z|Y)$ and $Q(z|X)$ can be used to evaluate the capability of the encoding network to generate latent variables which are likely to produce the expected target $Y$
\begin{align}
	\label{equ:target}
	& D[Q(z|X)||P(z|Y)]  \nonumber \\
	=& E_{z\sim Q}[\log Q(z|X) - \log P(Y|z) - \log P(z)] + \log P(Y).
\end{align}
When we expand the KL divergence in (\ref{equ:target}) to optimize $Q(z|X)$ based on the Bayes rule in (\ref{equ:bayesian1}),
$P(Y)$ still exists for calculating $P(z|Y)$ which is not tractable due to the same reason mentioned above.

So we rewrite (\ref{equ:target}) by combining the KL divergence with $\log P(Y)$ and represent them as the new optimization target $B(P,Q)$, which is the evidence lower bound of $P(Y)$, to get rid of the process for calculating $P(Y)$ directly for optimizing $Q(z|X)$ in divergence
\begin{align}
	\label{equ:lower_bound_meaning}
	B(P,Q) = \log P(Y) - D[Q(z|X)||P(z|Y)]~.
\end{align}
So our target becomes maximizing the evidence lower bound $B$ with respect to $\log P(Y)$ to make $D[Q(z|X)||P(z|Y)]$ as small as possible.
To implement our target via deep learning, we define the cost function $\mathcal{L}_\text{enc-gen} = -B$ of the generative model.

By expanding the divergence $D[Q(z|X)||P(z|Y)]$ in (\ref{equ:lower_bound_meaning}) and combining $\log Q(z|X)$ and $-\log P(z)$ together as another divergence, we can obtain the final form of the cost function as our encoder and decoder
\begin{align}
	\label{equ:loss_gmcml}
	\mathcal{L}_\text{enc-gen} =& E_{z\sim Q}[\log P(Y|z) + \log P(z) - \log Q(z|X)] \nonumber \\
	=& \underbrace{D[Q(z|X)||P(z)]}_\text{$\mathcal{L}_\text{enc}$} \underbrace{- E_{z\sim q}[\log P(Y|z)]}_\text{$\mathcal{L}_\text{gen}$}~.
\end{align}
The final loss function in (\ref{equ:loss_gmcml}) can be interpreted as follows: we replace the information component $\log P(X|z)$ used in VAEs as $\log P(Y|z)$ to make the decoding process suitable for different targets $Y$.

Traditionally, foreground object reconstruction is treated as pixel-wise segmentation using one-hot coding like DCN~\cite{Noh2015Learning}, so the number of output channel is no fewer than the number of categories of training data.
In our work, semantic reconstruction results are represented in RGB channels which could be easily applied on different databases because the number of channels are fixed as 3.

\section{Categorical Classification with Metric Learning\label{sec:classification}}

Object recognition is dependent on categorical information, so we concatenate a classification network after the generative sub-network.
Assuming that a descriptor of the reconstructed channels from one sample is represented as $r=[pixels_{r}, pixels_{g}, pixels_{b}]$ where $pixels$ are feature for a single channel,
the effectiveness of back-propagated Euclidean loss directly depends on the assigned absolute RGB values of masks which means that greater disparity between two classes matters more than the minor one.
This is a disaster when there are so many categories that we can't attribute effective RGB colors anymore for a discriminative rendering.

We make features of our training data and real photos similar to each other by exploiting common factors to minimize over-fitting problem triggered by absence of textures.
Based on additional information such as object poses, ZigzagNet is used as architecture for pre-training which is optimised by a multi-triplet cost function together with an optional pairwise term for speeding up convergence.
Then additional synthetic data with different focal points is used for fine-tuning with softmax loss.
Our training method makes the descriptors in classification sub-network suitable for Nearest Neighbour (NN) classifier without softmax and support vector machine classifier due to data migration problem between synthetic data and real images.
We evaluate the performance of our work on output of the penultimate layer of the classification network before the last inner product matrix.
Our discriminant descriptor satisfies NN classifier better than models trained from real images and is even similar to the performance of AlexNet~\cite{Krizhevsky2012} descriptors which is also trained from realistic data using larger parametric model.

\subsection{Compact Classification Architecture\label{sec:classification_compact_arch}}

With reconstruction sub-network 14 MB on disk,
we apply ZigzagNet for the conjugate classification which has 30$\times$ fewer parameters than AlexNet and is just 6.2 MB on disk to make the whole architecture as compact as possible.
So the overall model size of our conjugating model is just about 20.2 MB\@.
Over-fitting problem from synthetic data to real photos is also overcame by applying Zigzag module instead of simply using convolutional layer.

\begin{figure}[ht]
	\captionsetup[subfigure]{justification=centering}
	\centering
	\begin{subfigure}[t]{0.75\linewidth}
		\centerline{\includegraphics[width=1\linewidth]{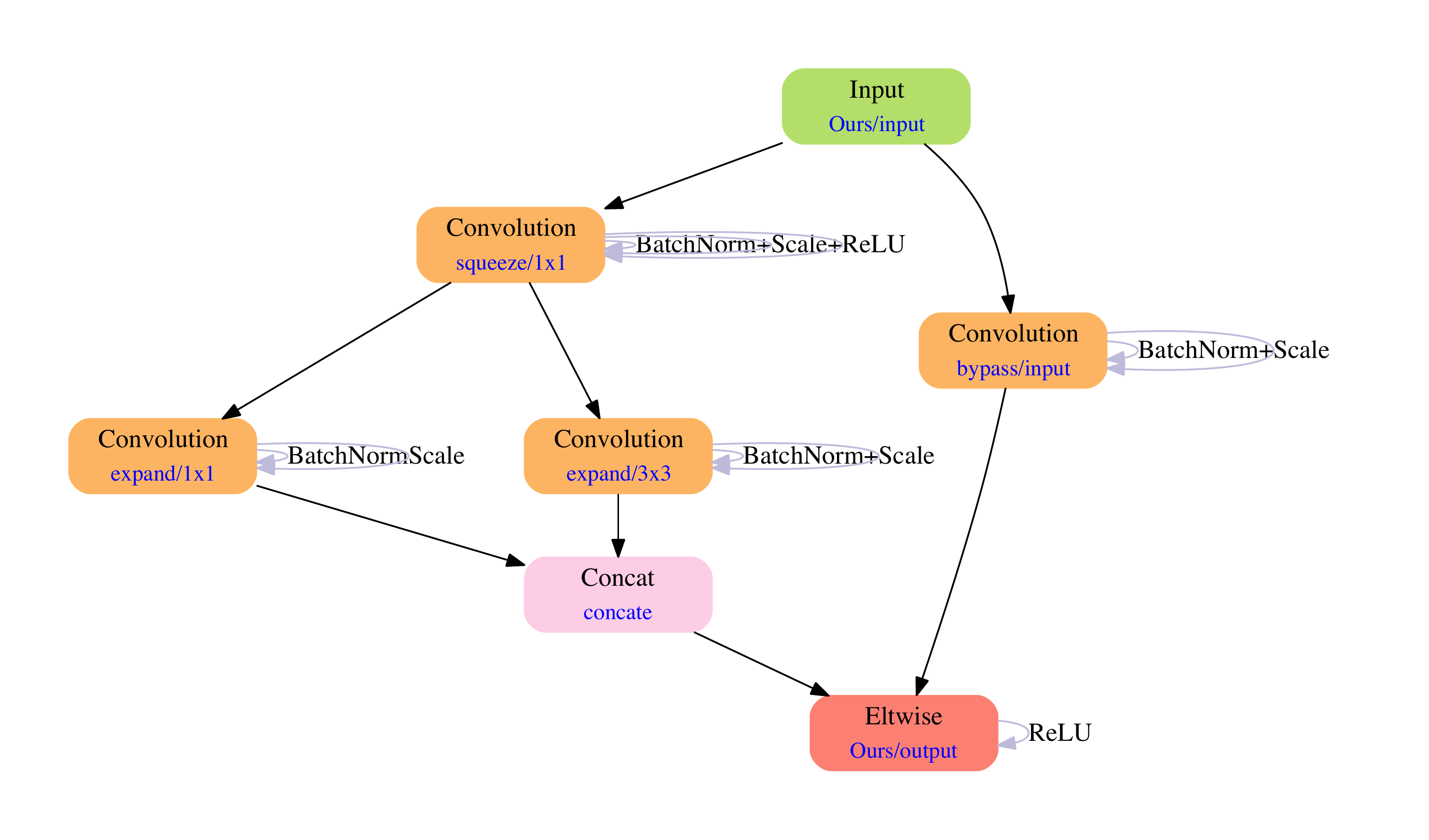}}
		\caption{Micro \label{fig:micro_zig}}
	\end{subfigure}
	\begin{subfigure}[t]{0.23\linewidth}
		\centerline{\includegraphics[width=1\linewidth]{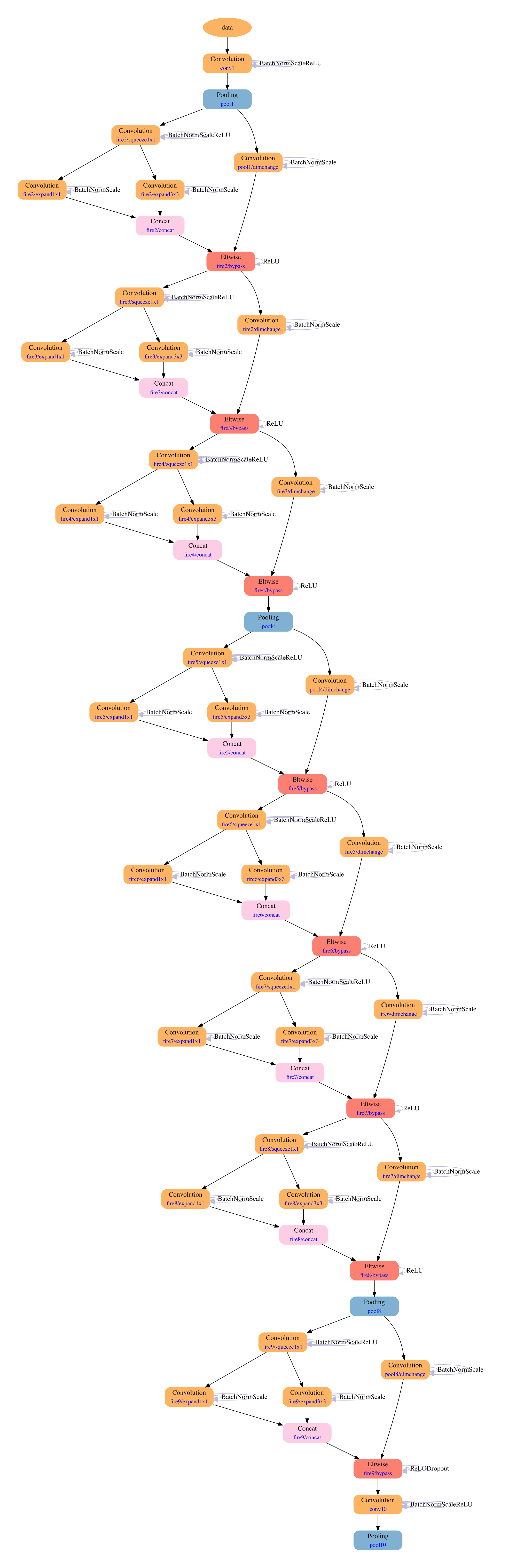}}
		\caption{Macro \label{fig:macro_ours}}
	\end{subfigure}
	\caption{Micro Zigzag module and macro ZigzagNet architecture.}
\end{figure}

We try to solve data migration problem between synthetic training data and real testing data by making the whole architecture more compact.
Due to the fact that we want to use the concatenated RGB channels and the generated object reconstruction channels which differ a lot directly as 6-channel input for classification sub-network,
so we adopt channel-wise compression between micro architectures for deep architecture to avoid over fitting problem.
We modify convolutional Fire module in SqueezeNet~\cite{IandolaMAHDK16} into Zigzag module to form ZigzagNet which performs better in classification compared to SqueezeNet.
ZigzagNet combines SqueezeNet with residual concept~\cite{he2015deep} without using any fully connected layers.
Macro architecture of ZigzagNet is concatenated of repetitive micro Zigzag modules composed of three $1\times1 $ and one $3\times3$ convolutional layers in a zigzag style of 3 steps: channel wise squeezing for principal representation, reception fields expanding by parallel connection of multi-scale convolution kernels and consolidation by adding $1\times1$ convolutional layer to bypass the original information before squeezing to keep the learning process stable while remaining the size of parametric model.
Input to a micro Zigzag module is compressed by a $1\times1 $ convolutional layer for channel-wise linear projection to make the following parameter model compact and representative.
This is pretty important for our data generated from reconstruction sub-network with concatenated synthetic reconstruction channels and realistic RGB channels because such $1\times1 $ convolutional operation behaves like a trainable preprocessing procedure before spatial convolutions afterwards.
Output in a single module within macro architecture is joined with bypassed input information by a convolutional layer as the input of next module to keep the learning process stable.
Our micro module also differs from Fire module of bypassed SqueezeNet shown in Fig.~\ref{fig:micro_zig} in nonlinear operation,
ReLU operation on the expanding layer before the element wise summation layer is moved to the output of Fire module to eliminate the scale difference between two inputs of the element wise operation layer and make information from both branches compressed at the same time.
For convenience of comparison experiments, depth of macro architecture of ZigzagNet is set as the same as AlexNet and SqueezeNet in account of micro architectures with channel-wise representation.
The final macro architecture of ZigzagNet is shown in Fig.~\ref{fig:macro_ours} which is a concatenated structure of several micro modules of Zigzag module shown in Fig.~\ref{fig:micro_zig}.

\subsection{Metric Learning with Multi-triplet Set\label{sec:classification_metric}}

Realistic images generated by two camera modes separately are both well utilised using multi-triplet cost and softmax cost.
As background images are attached behind the rendered objects for realistic images,
training with category labels based on texture-less models leads to over-fitting on distinguishing edges of objects against background.
If we just train classification model directly based on those synthetic data, objects without clear boundary against background in real images won't be recognized well.
We introduce a triplet cost to utilize information contained in poses, lighting conditions of synthetic data which are also common attributes for real photos.

\begin{figure}[ht]
	\captionsetup[subfigure]{justification=centering}
	\centering
	\begin{subfigure}[t]{0.8\linewidth}
		\centering
		\centerline{\includegraphics[width=1\linewidth]{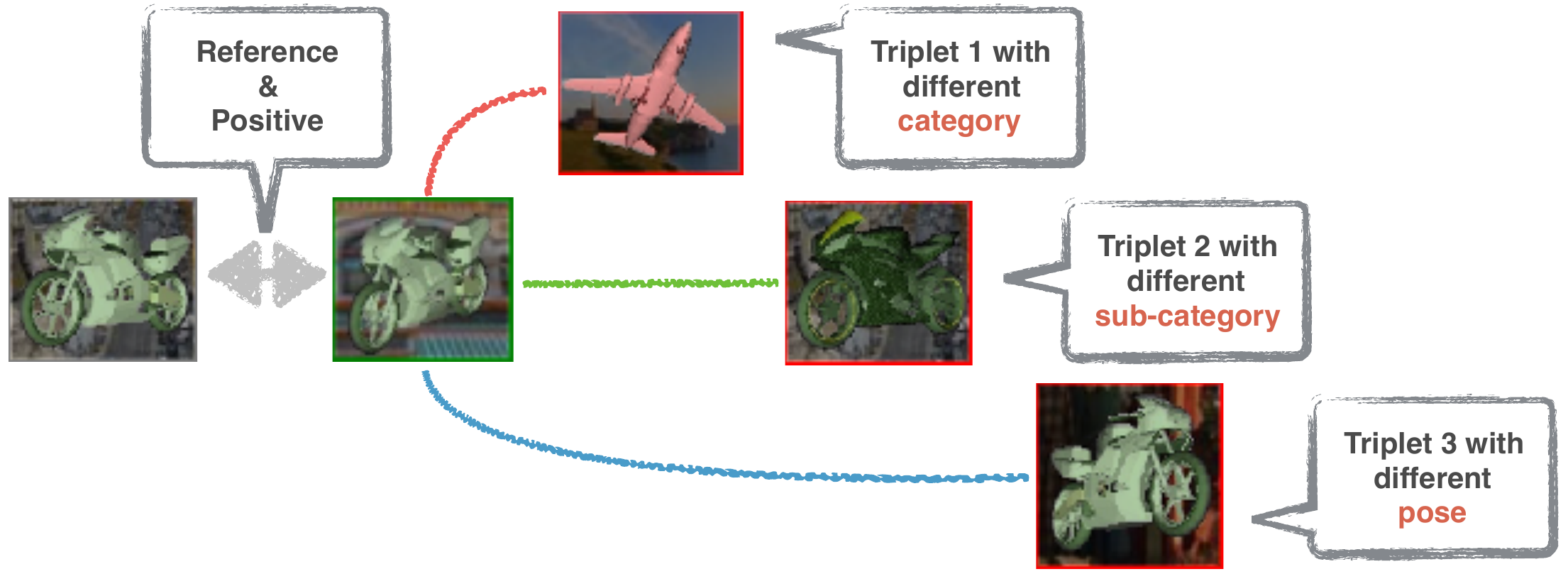}}
		\caption{Triplet set with 5 samples\label{tripletset}}
	\end{subfigure}
	\begin{subfigure}[t]{0.46\linewidth}
		\centering
		\centerline{\includegraphics[width=1\linewidth]{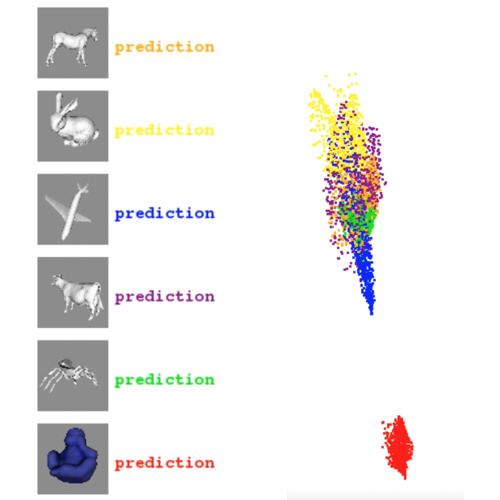}}
		\caption{Softmax cost distrbution\label{fig:feat_distri_model_soft}}
	\end{subfigure}
	\begin{subfigure}[t]{0.46\linewidth}
		\centering
		\centerline{\includegraphics[width=1\linewidth]{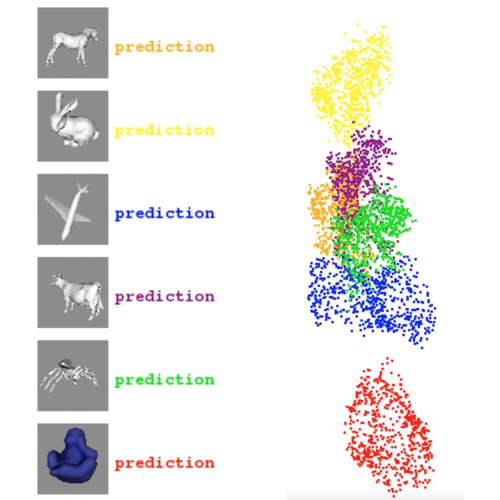}}
		\caption{Triplet cost distribution\label{fig:feat_distri_model_tri}}
	\end{subfigure}
	\begin{subfigure}[t]{0.46\linewidth}
		\centering
		\centerline{\includegraphics[width=1\linewidth]{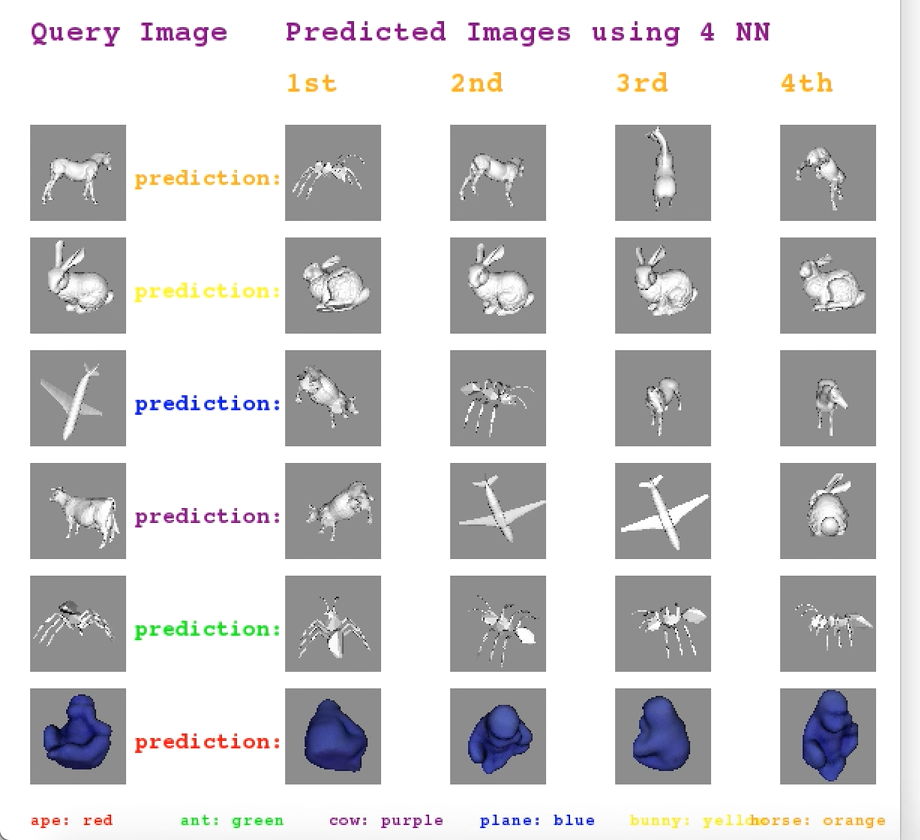}}
		\caption{Nearest Neighbor prediction on softmax training\label{fig:nn_softmax}}
	\end{subfigure}
	\begin{subfigure}[t]{0.46\linewidth}
		\centering
		\centerline{\includegraphics[width=1\linewidth]{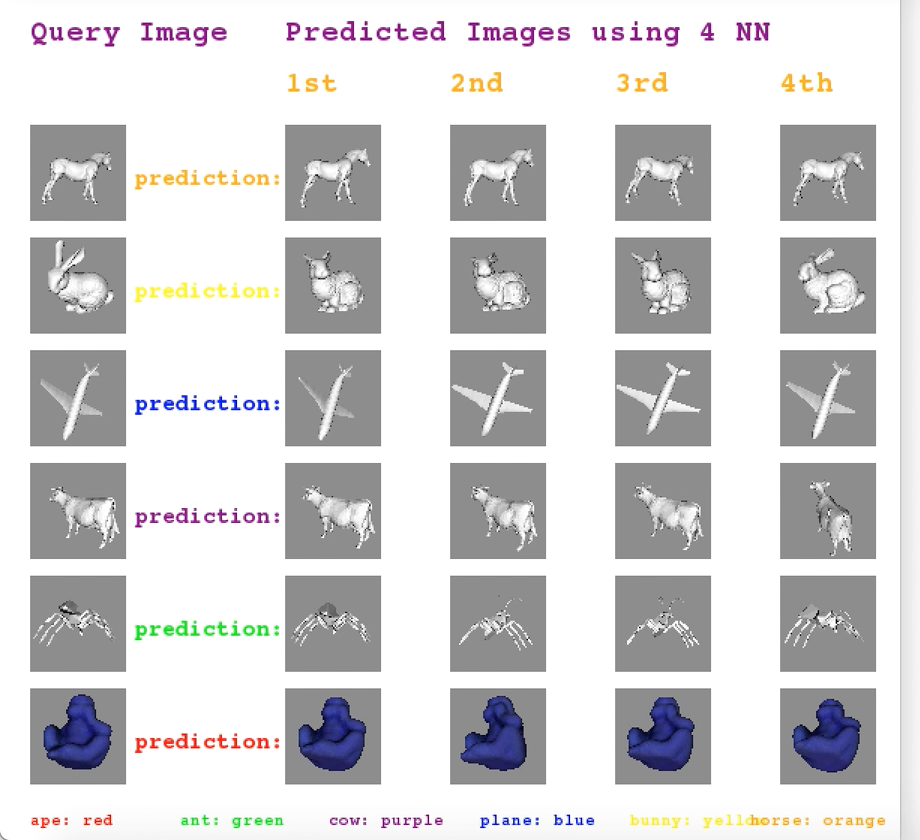}}
		\caption{Nearest Neighbor prediction on triplet training\label{fig:nn_triplet}}
	\end{subfigure}
	\caption{Triplet set arrangement and feature visualization with PCA matrix calculated from descriptors.\label{fig:feat_distri_model}}
\end{figure}

Our multi-triplet cost function which is modified from work of~\cite{wohlhart15} together with an optional pairwise term is used in the initial training process to improve the performance for recognizing physical characteristics of foreground objects no matter whether there is background or not.
Fig.~\ref{fig:feat_distri_model} shows that triplet training based on poses makes descriptors of synthetic images form in a sphere-like distribution and Nearest Neighbor Prediction with 4 candidates in Fig.~\ref{fig:nn_triplet} indicates that trained parametric model has ability on both identification on object categories and distribution regression on object poses.
Fig.~\ref{fig:feat_distri_model_soft} shows that training directly with softmax loss based on categorical information over fits the blue surface of ``Ape'' which is bad for distinguishing the other 5 categories with white surface.
Descriptors trained by our multi-triplet loss shown in Fig.~\ref{fig:feat_distri_model_tri} have a more reasonable distribution related to poses.
Triplet loss makes it possible to optimise the parametric model with common characteristics including object poses, lighting conditions and camera modes which are accurately recorded during rendering procedure of synthetic images.
Fully exploiting the information in poses can help to avoid learning bad relationships between objects and background which make those synthetic images not so realistic.
One multi-triplet set is composed of a reference sample, a positive sample and several negative samples:
\begin{align}
	\mathcal{L}_{triplet} = \mathcal{L}_{pair}(x_i,x_j)+\sum_{(x_i,x_j,x_k)\in{\mathcal{T}}} \mathcal{L}_{tri}(x_i,x_j,x_k)~,
\end{align}
where $ \mathcal{L}_{pair}(x_i,x_j) = ||f(x_i)-f(x_j)||_2^2 $.
As shown in Fig.~\ref{tripletset}, our multi-triplet set is composed of 5 samples,
the positive sample is the one with closest pose from reference sample or only has a different lighting condition which is the same pose and 3 negative samples are selected as one differs more in pose from the same class and two from other classes or sub-classes.
Our triplet loss makes the whole training process going smoothly and effectively by solving gradient vanishing and exploding.
Firstly, the problem of gradient vanishing in traditional triplet~\cite{WangSLRWPCW14} training where positive sample is similar to negative sample is solved by adopting basic form of triplet loss as the one in~\cite{wohlhart15}.
Feature distance is modified from Euclidean distance to its squared form to make distribution of learned descriptors has a manifold of sphere as shown in Fig.~\ref{fig:feat_distri_model_tri} which is related to geometric camera positions on a sphere.
This also makes the learning process more stable for 3 samples which differs little in a triplet set, so tiny difference between three samples without background with similar poses will not lead to a not-a-number gradient which might appear in~\cite{wohlhart15}.
Secondly, large variance in background images from Flickr makes the gradient explodes easily which is not so helpful for understanding hidden information of the foreground objects, so we further apply natural logarithms on loss of a triplet:
\begin{align}
	\mathcal{L}_{tri} = \ln(\max(1,2-\frac{||f(x_i)-f(x_k)||_2^2}{||f(x_i)-f(x_j)||_2^2 + m_{\text{tri}}}))~,
\end{align}
where $ f(x) $ is the input of the loss for sample $ x $ and $ m_{\text{tri}} $ is the margin for triplet.
Denote that $D_{ij}=||f(x_i)-f(x_j)||_2^2$ and $D_{ik}=||f(x_i)-f(x_k)||_2^2$,
so the partial differential equations for the input of triplet loss layer are:
\begin{align}
	\dfrac{\partial \mathcal{L}_{tri}}{\partial f(x_i)}=&\frac{D_{ik}(f(x_i)-f(x_j))-(D_{ij}+m_{\text{tri}})(f(x_i)-f(x_k))}{\mathcal{L}_{tri}{(D_{ij}+m_{\text{tri}})}^2} \nonumber \\
	\dfrac{\partial \mathcal{L}_{tri}}{\partial f(x_j)}=&\frac{D_{ik}(f(x_j)-f(x_i))}{\mathcal{L}_{tri}{(D_{ij}+m_{\text{tri}})}^2} \nonumber \\
	\dfrac{\partial \mathcal{L}_{tri}}{\partial f(x_k)}=&\frac{f(x_i)-f(x_k)}{\mathcal{L}_{tri}(D_{ij}+m_{\text{tri}})}
\end{align}
For convenience of an easier training data preparation procedure,
we set the positive and the reference samples fixed in a multi-triplet set and use them directly as the pairwise term if it is needed.
As training based on multi-triplet sets costs more time for convergence, samples without background are used at first stage to make parametric model converge faster.
The pairwise term is only applied to make the descriptor robust to small variance of objects rather than the more complicated backgrounds images which are added in the following training stage.

\subsection{Fine-tuning with Additional Data\label{sec:classification_tuning}}

Triplet training based on pose information uses synthetic samples where focal points are all in the center of every objects.
We add softmax loss for fine-tuning based on close shot images with shifting focal points.
Distribution of descriptors projected with 3 dimensional PCA matrix shows that special samples from real test images could be better clustered with other normal samples in the same class after fine-tuning and all samples are set apart better according to categories.
Additional rendered data gives a help on reducing intra-class variability while keeping the geometry relationship for descriptors of regular samples based on pre-trained model.

\section{Coordinate Training for Conjugate Structure\label{sec:coordinate}}

As shown in Fig.~\ref{fig:pipeline_tip}, the concatenated reconstruction and classification sub-networks both have inputs of realistic images which means that both sub-networks are relatively coordinate in training stage.
Here coordinate training means that output of reconstruction sub-network always help to optimize classification sub-network and the gradients back-propagated from the classification sub-network are always effective.
Our coordinate training uses coupled noising ratio $R_\text{rec}$ and $R_\text{cls}$ to corrupt realistic inputs for the aim of feeding effective information back and forth between two sub-networks.
The following part introduces problems in concatenated network and introduce how our coupled noise ratios on each input of sub-networks help to efficiently train both sub-networks.

\subsection{Adaptive Noise Assistant for Information Feeding\label{sec:coordinate_noise}}

We makes two sub-networks trained more effectively by applying adaptive noises for realistic input channels.
Two sub-networks should not be simply concatenated together because different convergence speed of sub-networks will make it hard training the second network according to the output of the first network and utilizing the back-propagated gradients for the first network.
As shown in Fig.~\ref{fig:rec_corrup_noise}, reconstruction sub-network is only able to reconstruct average images of all semantic masks in the beginning,
so these three additional channels are not so discriminant that the classification network can not utilize them compared to the realistic channels.
Assume that synthetic training data is evenly distributed in categories and two types of images are concatenated together as inputs for classification sub-network,
the classification results will be similar no matter whether using the reconstruction network as supporting architecture or not.
The variance between original images and reconstructed images are so large that the statistically back-propagated gradients could not be effectively passed back to the reconstructed channels according to loss functions designed for classification network.
We present synthetic images as object channels $O$, semantic depth masks as mask channels $M$, reconstruction network with input $I_{\text{rec}}=[O]$ produces reconstruction channels $M'$ similar to $M$ and the data fed in classification network is represented as a channel-wise concatenated tensor as $I_{\text{cls}}=[O, M']$.
Direct training for classification network based on $[O, M']$ will lead to a severe problem of over-fitting on reconstructed channels because the variance of $M'$ is far smaller than the other three object channels which means that the classification network utilizes few information from reconstructed channels,
so the classification result differs little even without $M'$ which means that $cls(I_{\text{cls}}) \approx cls(I_{\text{rec}})$.
For perspective of statistical gradient decent for the reconstruction sub-network, gradients feeding back through $ I_{\text{cls}} $ are not effective due to the absence of correlation between the classification loss and information in $ I_{\text{cls}} $.
This means that the two parts of the whole network are not so relevant to each other and also can not take benefits from each other's forward outputs and back-propagated gradients.

We propose an efficient strategy using adaptive noise on the realistic data separately on the input of reconstruction network and classification network to solve this problem.
We utilize coupled noising function $n_{\text{rec}}$ and $n_{\text{cls}}$ for inputs of reconstruction network $I_{\text{rec}}$ and classification network $I_{\text{cls}}$ to restrict information feeding in the original realistic images separately which are defined as
\begin{align}
	n_{\text{rec}}(I_{\text{rec}}) = & R_\text{rec}\ Ran + (1 - R_\text{rec})\ I_{\text{rec}} \nonumber \\
	n_{\text{cls}}(I_{\text{cls}}) = & R_\text{cls}\ Ran + (1 - R_\text{cls})\ I_{\text{cls}}~,
\end{align}
where $Ran$ is Gaussian noise, $R_\text{rec}$ is noising ratio for synthetic input data fed in reconstruction sub-network and $R_\text{cls}$ is noising ratio for classification sub-network.
Here we ensure that information from the realistic synthetic images are partially utilized according to the variation of the reconstructed samples and ground truth semantic masks.
Noising ratio $R_\text{rec}$ is set as result of hyperbolic tangent function applied on ratio of variance between reconstructed images and semantic masks in current batch while noising ratio $R_\text{cls}$ is obtained by hyperbolic tangent function applied on the result of $1 - \gamma(\alpha, \beta)R_\text{rec}$,
so we can see that $R_\text{rec}$ and $R_\text{cls}$ are strongly correlated with each other through $\gamma(\alpha, \beta) $.
Assuming that hyper parameters $\alpha$ and $\beta$ are set to be the same,
$\gamma $ will be 1 which means that we can easily represent $R_\text{cls}$ using $R_\text{rec}$.
Both ratios could be represented in (\ref{equ:noise_ratios}).
\begin{align}
	\label{equ:noise_ratios}
	R_\text{rec} = & \tanh(\alpha \frac{Var(M')}{Var(M) + m_{\text{noise}}}) \nonumber \\
	R_\text{cls} = & \tanh(1 - \tanh(\beta \frac{Var(M')}{Var(M) + m_{\text{noise}}}))
\end{align}

Here we usually set hyper parameters $(\alpha, \beta)$ as 0.25 and 2 according to results of experimental tests.
Generally speaking, the noising ratio for classification network $R_\text{cls}$ and the one for reconstruction network $R_\text{rec}$ are coupled as (\ref{equ:ralation_noise}).
\begin{align}
	\label{equ:ralation_noise}
	R_\text{cls} = \tanh(1 - \tanh(\frac{\beta}{\alpha}arctanh(R_\text{rec})))
\end{align}

Such strategy on training procedure ensures that information in realistic synthetic images is well utilized together with reconstructed images because classification sub-network tries best to learn from reconstructed channels without too much discriminant information in image channels in the beginning,
meanwhile the orderly concatenated sub-networks and spatially concatenated six-channel data which is the junction between two sub-networks are trained in balance with effective gradients.
Our coordinate training strategy makes the whole network converges easier than generative adversarial networks (GAN)~\cite{NIPS2014_5423, DBLP:journals/corr/RadfordMC15} where the generator and discriminator are sometimes trained separately according to the loss or discriminative accuracy,
our method could also utilize metric learning to do multi-task learning compared to GAN with a flexible classification network.
Similarly, parametric model in our network benefits from each others when foreground object reconstruction and classifier are always supervising each others in the training stage.
If we regard them as generator and discriminator like concept in GAN,
the parametric model are updated with different paces which means that it's always updating parametric model even with zero gradients defined by the variance of generated channels.

\section{Overall model and implementation\label{sec:model}}


The overall loss function for our model can be thus defined as:
\begin{align}
	\label{equ:loss_total}
	\mathcal{L}_{total} = \mathcal{L}_\text{enc-gen} + \mathcal{L}_\text{pair} + \mathcal{L}_\text{tri} + \mathcal{L}_\text{s}~,
\end{align}
The variational prediction loss ($\mathcal{L}_\text{enc-gen}$) is summed with the pair-wise loss ($\mathcal{L}_\text{pair}$) and triplet loss ($\mathcal{L}_\text{tri}$). If we want to do classification for real photos, then the softmax loss ($\mathcal{L}_\text{s}$) is also added.

Similar to optimization process of VAEs~\cite{Kingma2013}, since the probability density function of the encoder, $Q(z|X)$, is still restricted to being a normal distribution $N(z|\mu$,$\Sigma)$, where $\mu(X;\theta)$ and $\Sigma(X;\theta)$ are arbitrary deterministic functions.
The formulation of the KL divergence that we employ is:
\begin{align}
	&D[N(\mu(p),\Sigma(p))||N(0,I)] \nonumber \\
	=& \frac{1}{2}{(\textrm{tr}{(\Sigma{(p)})}+{(\mu{(p)})}^T\mu{(p)}-k-\log \det {(\Sigma{(p)})})}~,
\end{align}
where $\textrm{tr}()$ is the trace and $\det()$ is the determinant. Here $\Sigma(p)$ is represented as $\log \Sigma(p)$.

One sample of $z$ for $P(Y|z)$ could be an approximation of $E_{z\sim Q}$ $[\log P(Y|z)]$. So, the full equation to be optimized is
\begin{align}
	\label{equ:optimize}
	E_{X}[E_{z\sim Q}[\log P(Y|z)] -D[Q(z|X)||P(z)]]~.
\end{align}
The gradient symbol of this equation can be moved into the expectations. Therefore, we can sample values of $z^{(l)}$ from the distribution $q(z|X)$, and compute the gradient of
\begin{align}
	\label{equ:gradient}
	\mathcal{L}_{r} =\frac{1}{L}\sum_{l=1}^{L}\log P(Y|z^{(l)})-D[Q(z|X)||P(z)]~.
\end{align}
We can then average the gradient of this function over samples of $X$ and $z$, so the result converges to the gradient of (\ref{equ:optimize}).
The problem in (\ref{equ:optimize}) is that $E_{z\sim Q}[\log P(Y|z)]$ depends on both parameters $p$ and $q$. We solve this problem in (\ref{equ:gradient}) by means of the re-parameterization trick used for VAE and explained in~\cite{Kingma2013}.

In conclusion, our coordinate training makes training process more efficient with one loss functions of (\ref{equ:loss_total}) for concatenated architectures.
In functional representation of our deep architecture, the whole parametric model is tuned in the same time.
	Suppose that we have paired synthetic training data of realistic images $O$ and semantic depth masks $M$, reconstruction network $rec()$ with mapping network $map()$ and rendering network $ren()$, classification network $cls()$ lead by multi-triplet cost function, paired adaptive noising $noi()$ and $noi'()$, the concatenated network could be represented as $cls(ren(map(noi(O))),\ noi'(O))$ in training stage and $cls(ren(map(O)),\ O)$ in testing stage.
	The generative model is embedded in the mapper $map()$ and render $ren()$ while metric learning is applied in the classifier $cls()$.
In the end, we make all those functions work as a whole architecture in perspective of effective parametric model training with paired noising functions $noi()$ and $noi'()$.

\section{Experiments}

\subsection{Experiments on ShapeNet Database}

\subsubsection{Visualization of Reconstruction}

\begin{figure}[ht]
	\captionsetup[subfigure]{justification=centering}
	\centering
	\begin{subfigure}[t]{0.24\linewidth}
		\centering
		\centerline{\includegraphics[width=1\linewidth]{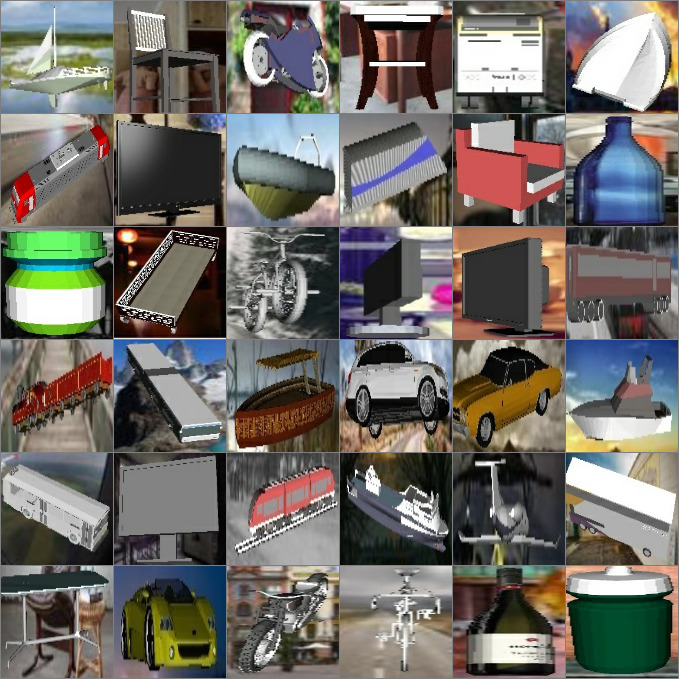}}
		\caption{input\label{fig:recon_train_raw}}
	\end{subfigure}
	\begin{subfigure}[t]{0.24\linewidth}
		\centering
		\centerline{\includegraphics[width=1\linewidth]{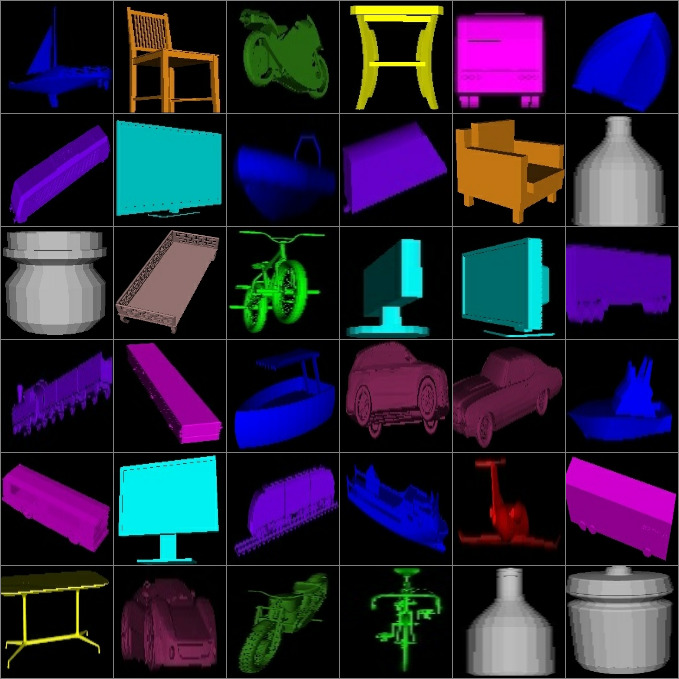}}
		\caption{target\label{fig:recon_train_mask}}
	\end{subfigure}
	\begin{subfigure}[t]{0.24\linewidth}
		\centering
		\centerline{\includegraphics[width=1\linewidth]{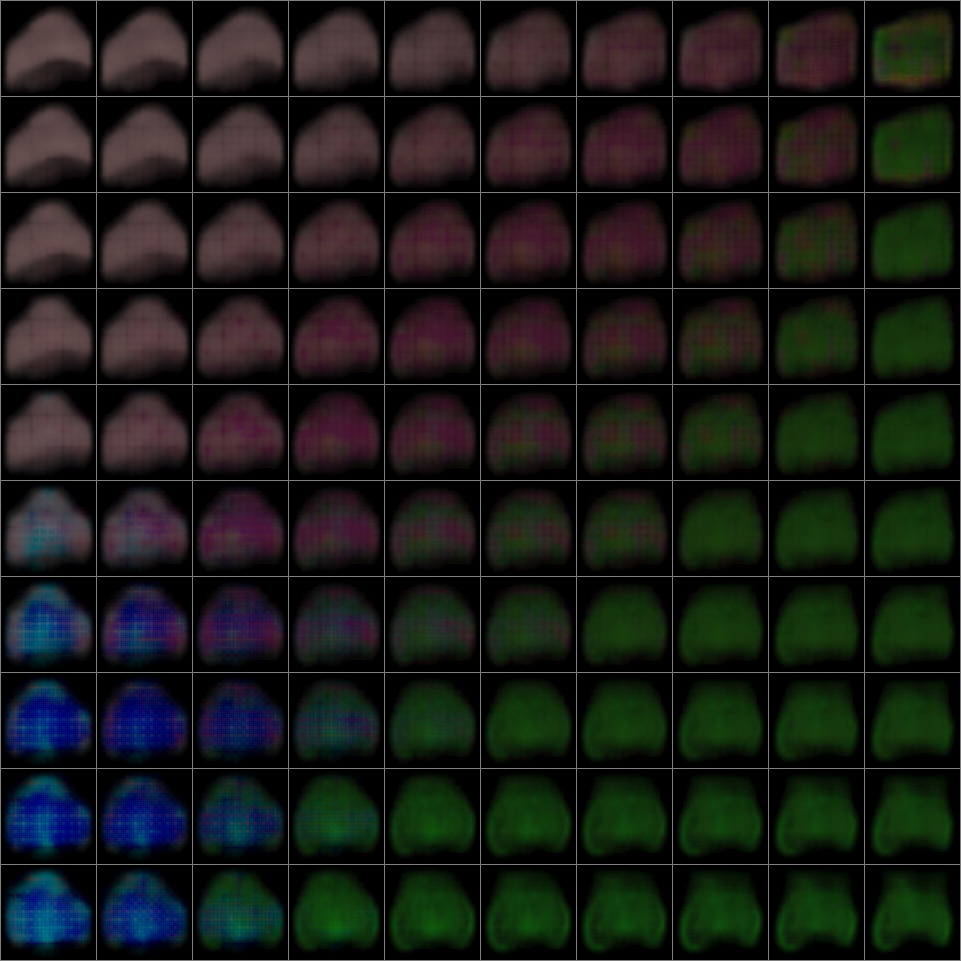}}
		\caption{visualisation of manifold\label{fig:recon_train_man}}
	\end{subfigure}
	\begin{subfigure}[t]{0.24\linewidth}
		\centering
		\centerline{\includegraphics[width=1\linewidth]{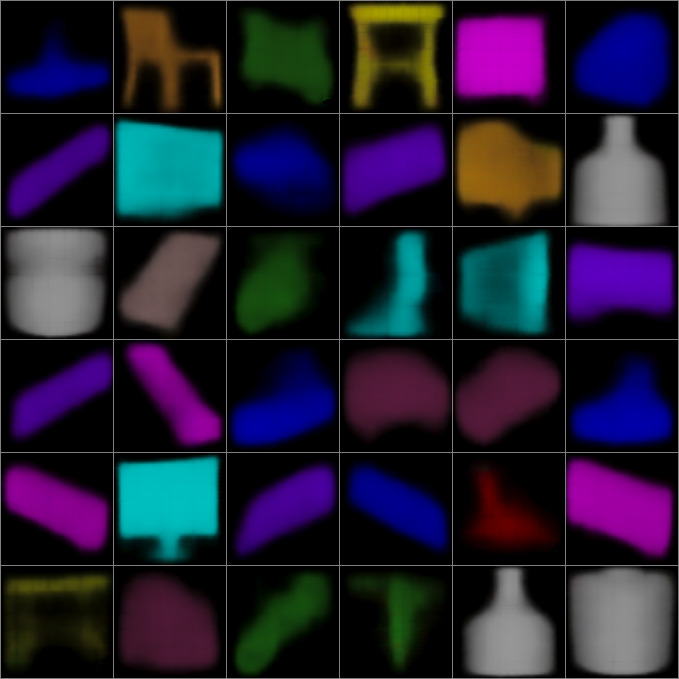}}
		\caption{output\label{fig:recon_train_recon}}
	\end{subfigure}
	\caption{Experiments on synthetic training data from ShapeNet models.\label{fig:target_shapenet}}
\end{figure}

Both training and testing data are synthetic realistic images with little data migration problem rendered from 3D models in 12 categories provided by ShapeNet for basic experiments to prove that our generative model has a strong fitting capability and semantic embeddings.
Reconstruction results of synthetic training images rendered from ShapeNet database show that our architecture is able to reconstruct ideal pixel-wise semantic foreground objects with expected colors.
Objects with 12 unique RGB colors representing for different classes are used as semantic reconstruction targets for training shown in Fig.~\ref{fig:recon_train_mask} which means that the reconstructed channels are represented by RGB values rather than channel-wise one-hot coding mask~\cite{Kheirandish2012} used in AAE~\cite{makhzani2015adversarial} with unfixed number of channels.
There are two main advantages contributed by reconstruction using RGB channels.
Firstly, reconstructed RGB channels could be widely used for representing segmentation masks in real database because itself is a kind of visualisation.
Secondly, semantic segmentation masks represented using one-hot coding have large amount of channels which make the pixel-wise representation so sparse that can hardly be stored efficiently on disks.
Our reconstruction cost function defined using Euclidean distance suits for optimising the decoder of the variational generative model.
It does not have dependency on uniform distribution on three RGB channels which means that semantic masks in two different categories with similar colors are not less discriminative compared to ones in another pair of colors with bigger Euclidean distance.
Our variational deep architecture is more discriminant and robuster in perspective of the reconstruction result.
Reconstruction target in form of semantic depth information shown in Fig.~\ref{fig:recon_train_mask} could be retrieved well in Fig.~\ref{fig:recon_train_recon} while suppressing the background at the same time.
Fig.~\ref{fig:recon_train_man} is the outputs of latent variables sampled in specific continuous range which shows that our generative reconstruction sub-network contains rich information about monotone surface colors of foreground object without background.
Similar architecture without variational inference is simply represented as FCN~\cite{Long2015Fully} in Fig.~\ref{fig:rec_fcn},
it shows that contours are not reconstructed as clear as our generative method and the colors in single object also differ more which means that there are not enough meaningful semantic information according to the categories of foreground objects.

As for the training process, reconstruction procedure shown in Fig.~\ref{fig:rec_corrup_noise} satisfies for our expectation on stability.
In initial training stage, the reconstruction sub-network is only able to reveal average color of all masks together with convex hull of objects by suppressing surrounding background.
Then the outputs of the reconstruction network according to input samples in categories with less intraclass variance such as bottles are reconstructed clearly.
Finally, objects are extracted from background together with relative depth information for samples in categories with larger intraclass variance like planes and bikes.

For analyzing how cost functions optimizing the reconstruction sub-network, visualizations of reconstructed channels during the whole training process are shown sequentially in Fig.~\ref{fig:rec_corrup_noise}.
The average color and contour of samples are reconstructed quickly in this initial stage.
In this stage, loss of generator $\mathcal{L_{\text{gen}}}$ based on Euclidean distance shown in Fig.~\ref{fig:linechart:lossnorm2} drops faster than Loss of encoder $\mathcal{L_{\text{enc}}}$ represented as KL divergence shown in Fig.\ref{fig:linechart:losscvae} which means that the reconstruction sub-network behaves similarly to reconstruction based on principle component analysis~\cite{Jolliffe1986Principal, Abdi2010Principal} to ensure reconstruction error as small as possible.
This means that samples in categories with little intraclass variance is reconstructed better in this stage.
Then encoder loss of KL divergence $\mathcal{L_{\text{enc}}}$ takes effect for rendering details to make sure that input samples from different categories with larger intraclass and smaller interclass variance can be reconstructed better contributed to latent variables.
Two components of the reconstruction loss play their own roles at different stages of training which means that $\mathcal{L_{\text{gen}}}$ make the generator quickly fit the mean value of the training data and simple samples,
then KL divergence component makes a finer fitting for interclass variance of samples while keeping previous samples reconstructed stably by utilizing latent space well with encoder.

\subsubsection{Semantic Embeddings Trained by Triplet Sets}

\begin{figure}[ht]
\captionsetup[subfigure]{justification=centering}
	 \centering
   \begin{subfigure}[t]{0.31\linewidth}
   \centerline{\includegraphics[width=1\linewidth]{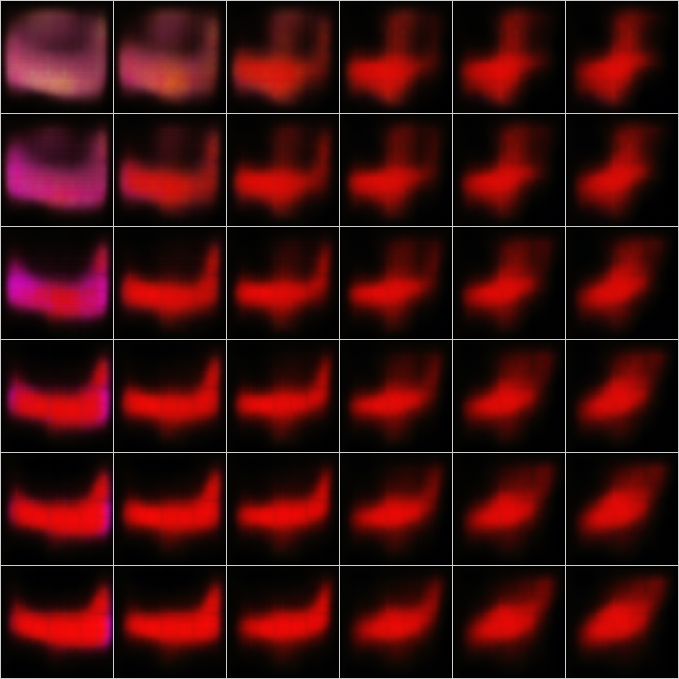}}
   \caption{plane}
   \end{subfigure}
   \begin{subfigure}[t]{0.31\linewidth}
   \centerline{\includegraphics[width=1\linewidth]{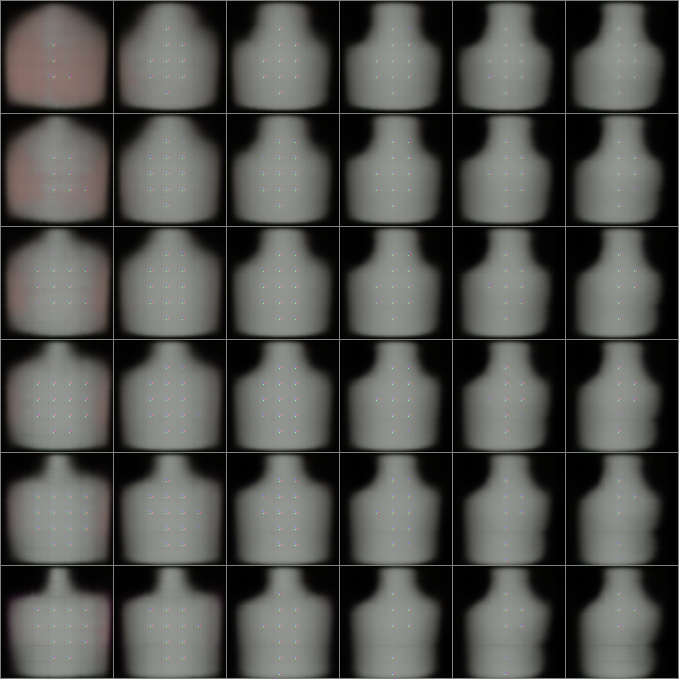}}
   \caption{bottle}
   \end{subfigure}
   \begin{subfigure}[t]{0.31\linewidth}
   \centerline{\includegraphics[width=1\linewidth]{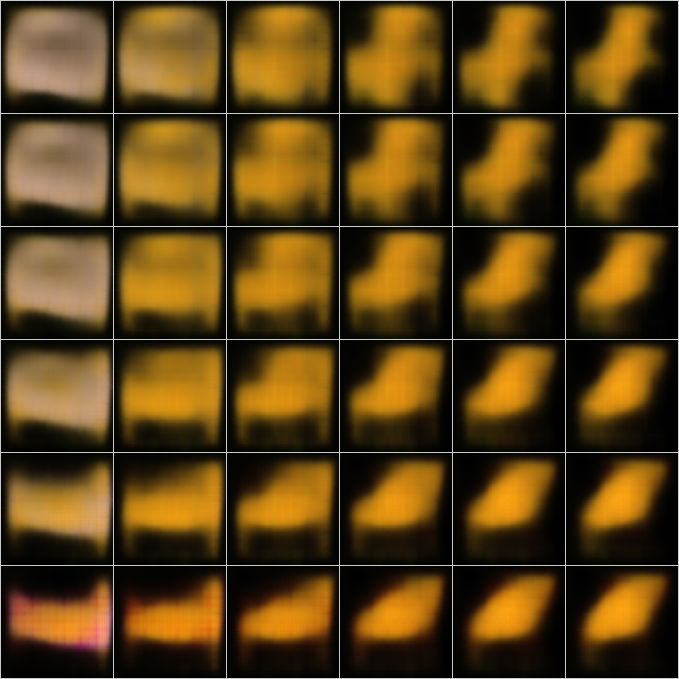}}
   \caption{chair}
   \end{subfigure}

   \begin{subfigure}[t]{0.31\linewidth}
   \centerline{\includegraphics[width=1\linewidth]{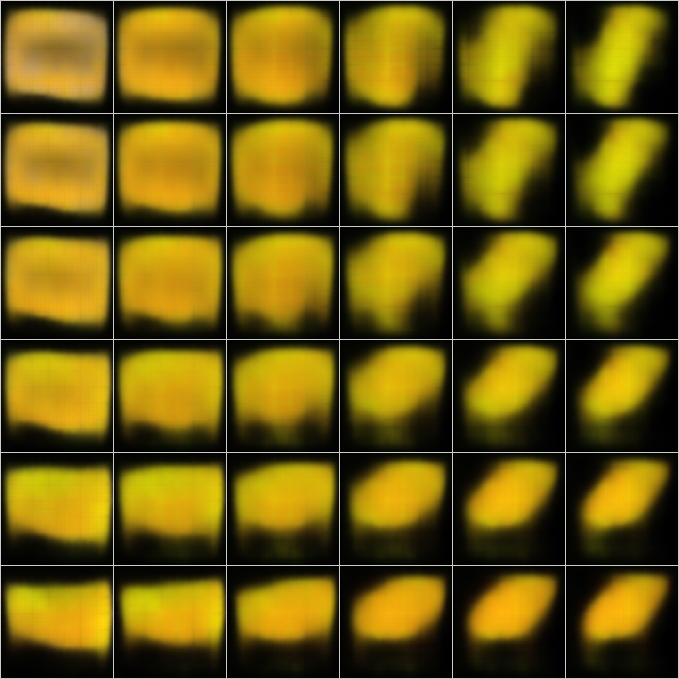}}
   \caption{table}
   \end{subfigure}
   \begin{subfigure}[t]{0.31\linewidth}
   \centerline{\includegraphics[width=1\linewidth]{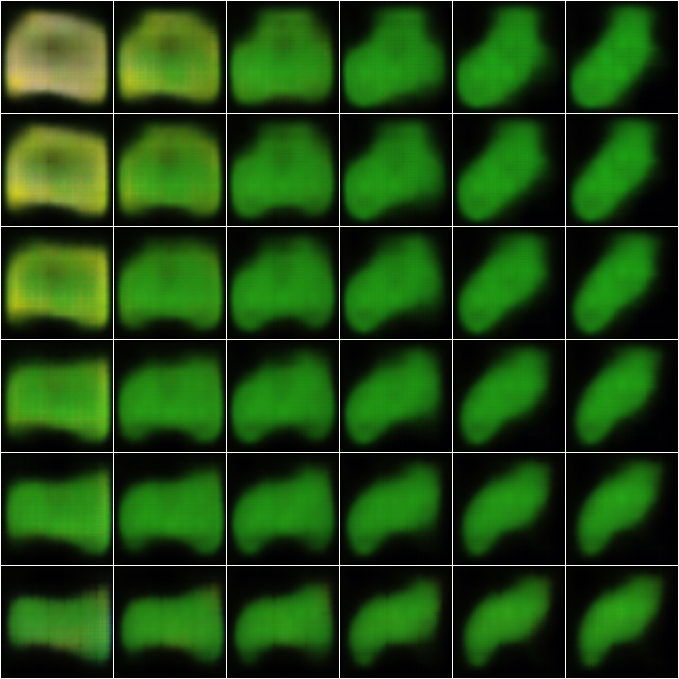}}
   \caption{motorbike}
   \end{subfigure}
   \begin{subfigure}[t]{0.31\linewidth}
   \centerline{\includegraphics[width=1\linewidth]{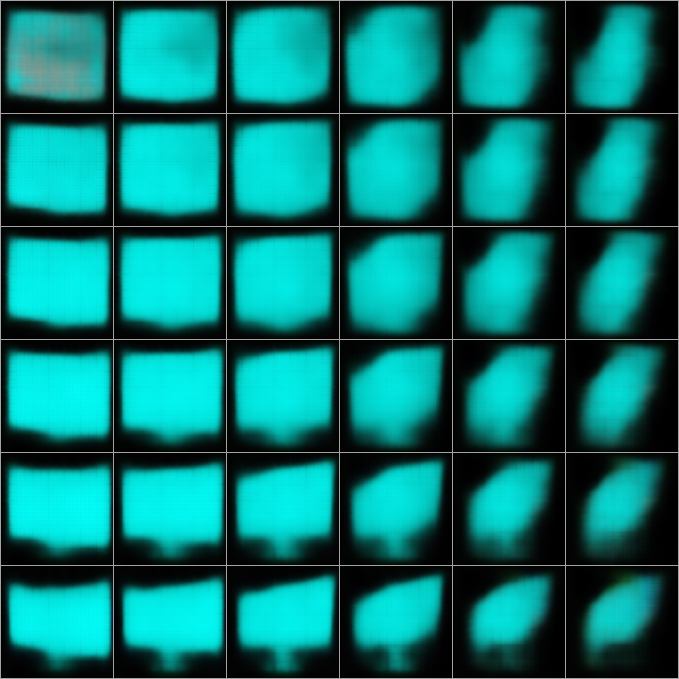}}
   \caption{monitor}
   \end{subfigure}
\caption{Visualized outputs by sampling cluster centers of latent variables from synthetic images.}
\label{fig:manifold_shapenet}
\end{figure}

For semantic rendering task shown in Fig.~\ref{fig:target_shapenet}, there are specific colors for each kind of objects in 12 categories.
As categorical information is included for setting up triplet sets, triplet training could also be applied on latent variables.
In such situation, clusters of latent variables from every categories could be set apart from each others.
Fig.~\ref{fig:manifold_shapenet} shows the visualization of output from latent samples $z$ of our reconstruction sub-network around cluster centers of every categories.
In each manifold, the poses of visualized outputs are continuous according to latent variables in specific range.
It is obvious that all of the rightmost samples at the bottom line have the same pose which means that our multi-triplet cost function helps to establish hidden semantic embeddings.

\subsubsection{Analysis for Optimizers in Training}

\begin{figure}[ht]
	\captionsetup[subfigure]{justification=centering}
	\centering
	\begin{subfigure}[t]{0.47\linewidth}
		\centering
		\centerline{\includegraphics[width=1\linewidth]{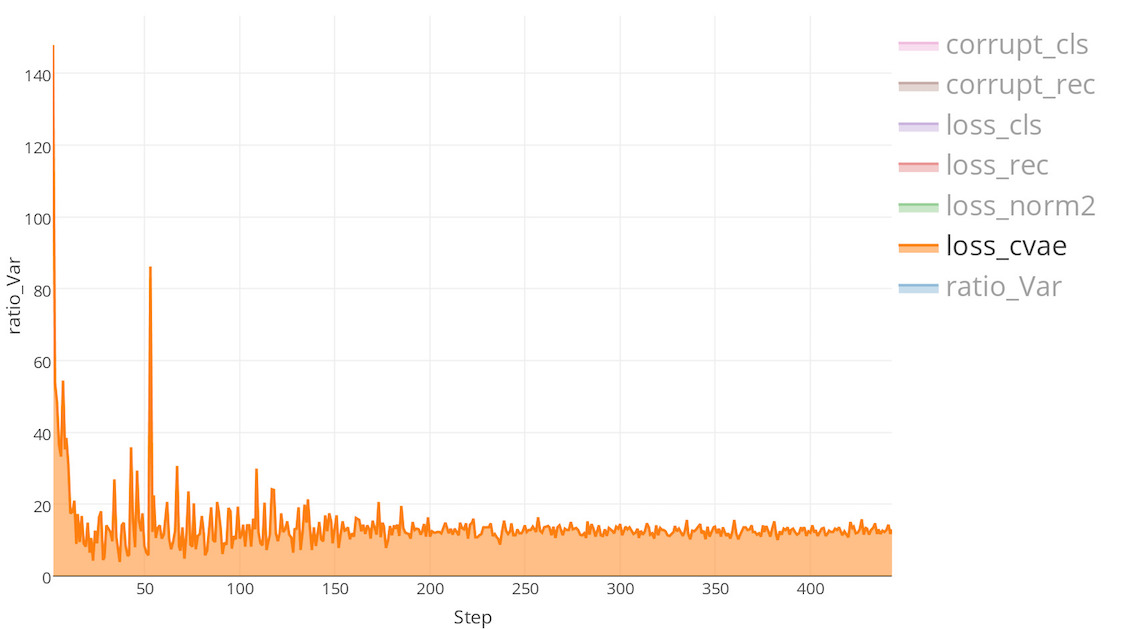}}
		\caption{KL divergence in reconstruction network\label{fig:linechart:losscvae}}
	\end{subfigure}
	\begin{subfigure}[t]{0.47\linewidth}
		\centering
		\centerline{\includegraphics[width=1\linewidth]{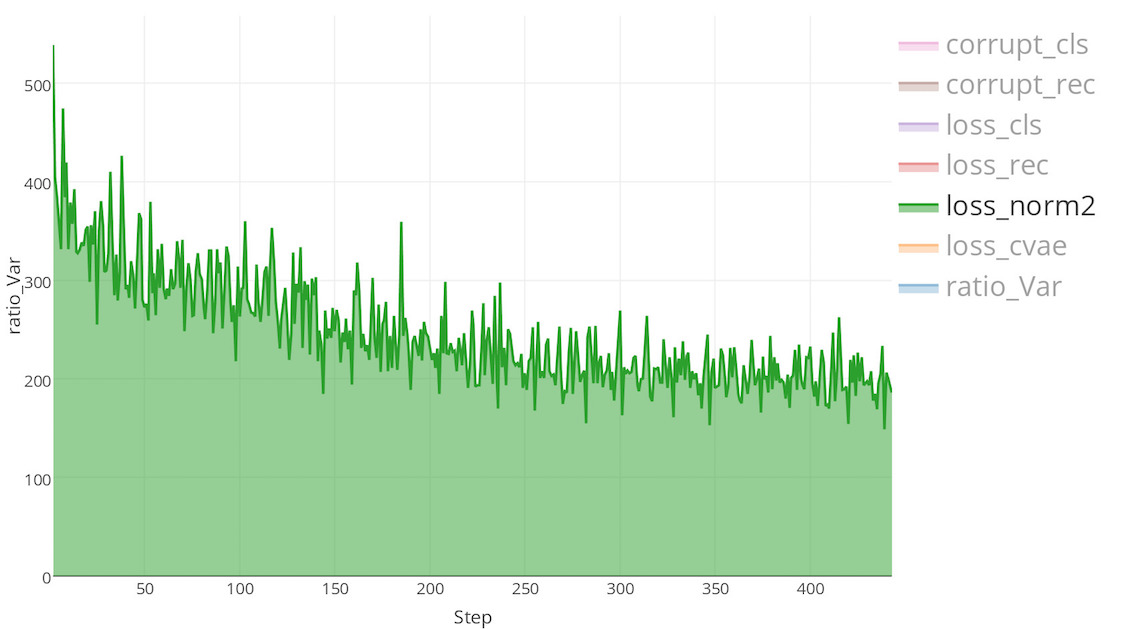}}
		\caption{Euclidean loss in reconstruction network\label{fig:linechart:lossnorm2}}
	\end{subfigure}

	\begin{subfigure}[t]{0.47\linewidth}
		\centering
		\centerline{\includegraphics[width=1\linewidth]{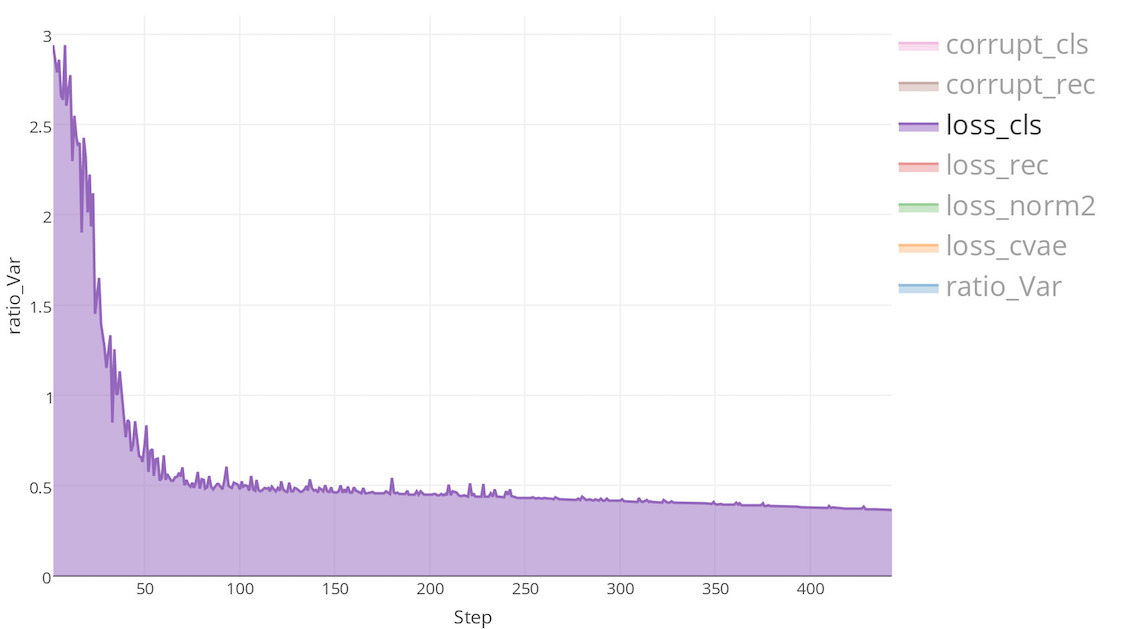}}
		\caption{classification loss\label{fig:linechart:losscls}}
	\end{subfigure}
	\begin{subfigure}[t]{0.47\linewidth}
		\centering
		\centerline{\includegraphics[width=1\linewidth]{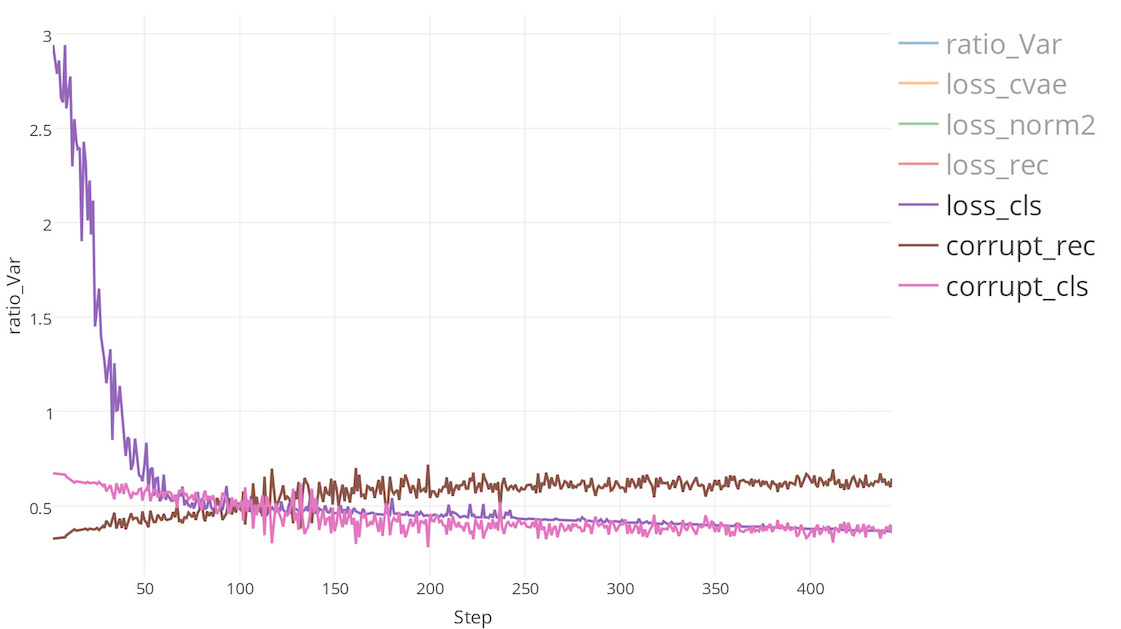}}
		\caption{classification loss and noising ratios\label{fig:linechart:losscls_corruprec_corrupcls}}
	\end{subfigure}

	\begin{subfigure}[t]{0.47\linewidth}
		\centering
		\centerline{\includegraphics[width=1\linewidth]{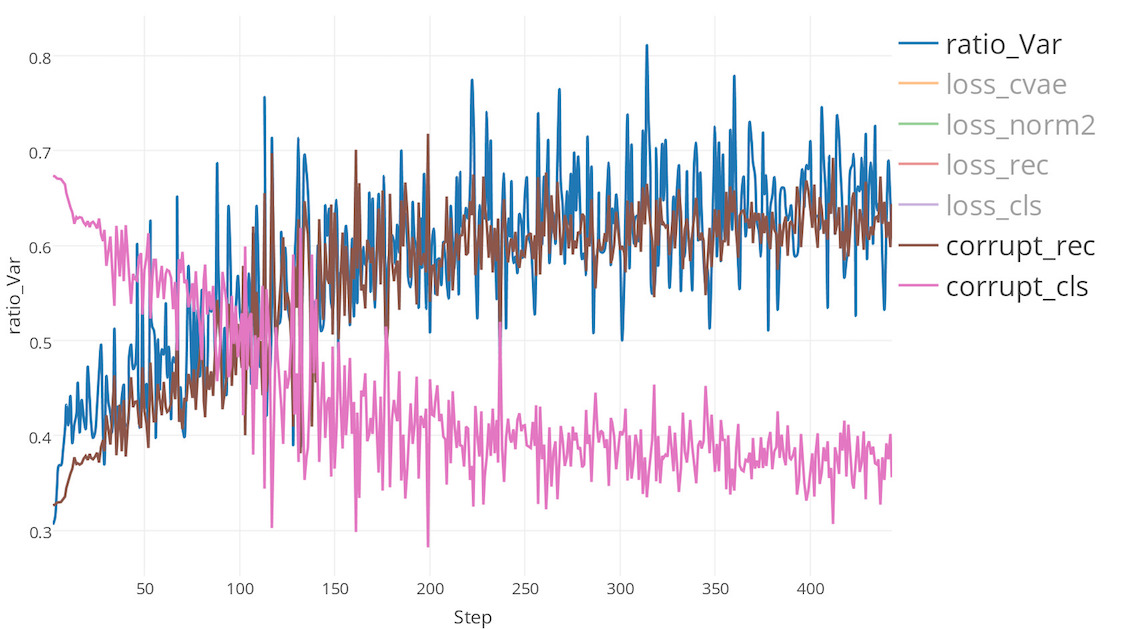}}
		\caption{noise and variation ratios\label{fig:linechart:ratio_corruprec_corrupcls}}
	\end{subfigure}
	\begin{subfigure}[t]{0.47\linewidth}
		\centering
		\centerline{\includegraphics[width=1\linewidth]{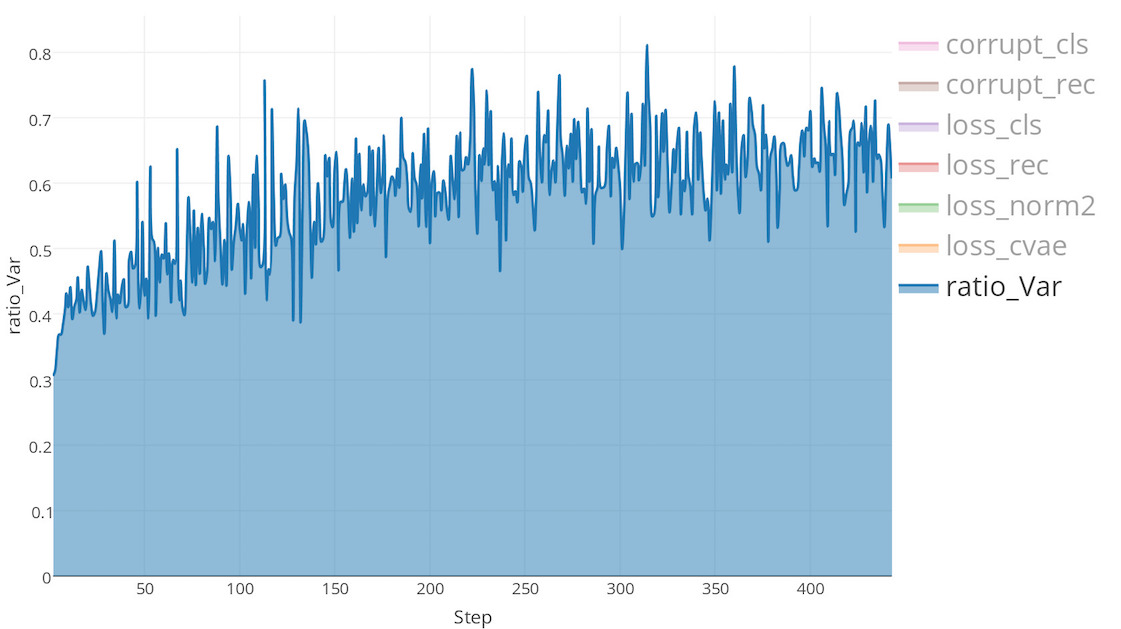}}
		\caption{variation ratio\label{fig:linechart:ratio}}
	\end{subfigure}

	\begin{subfigure}[t]{0.47\linewidth}
		\centering
		\centerline{\includegraphics[width=1\linewidth]{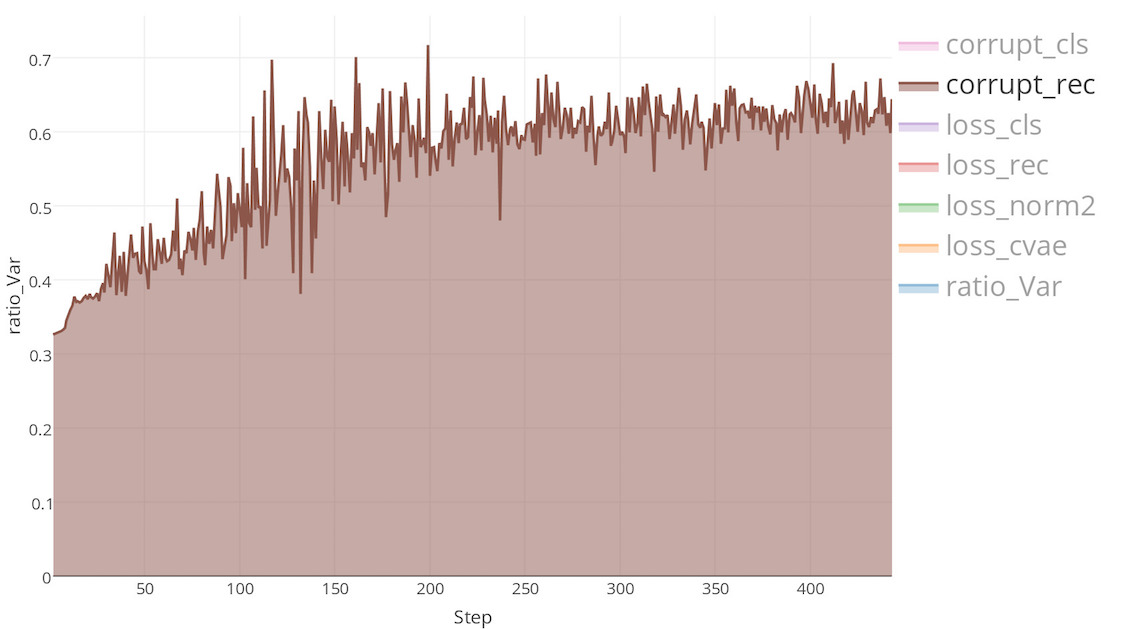}}
		\caption{reconstruction noise ratio\label{fig:linechart:corruprec}}
	\end{subfigure}
	\begin{subfigure}[t]{0.47\linewidth}
		\centering
		\centerline{\includegraphics[width=1\linewidth]{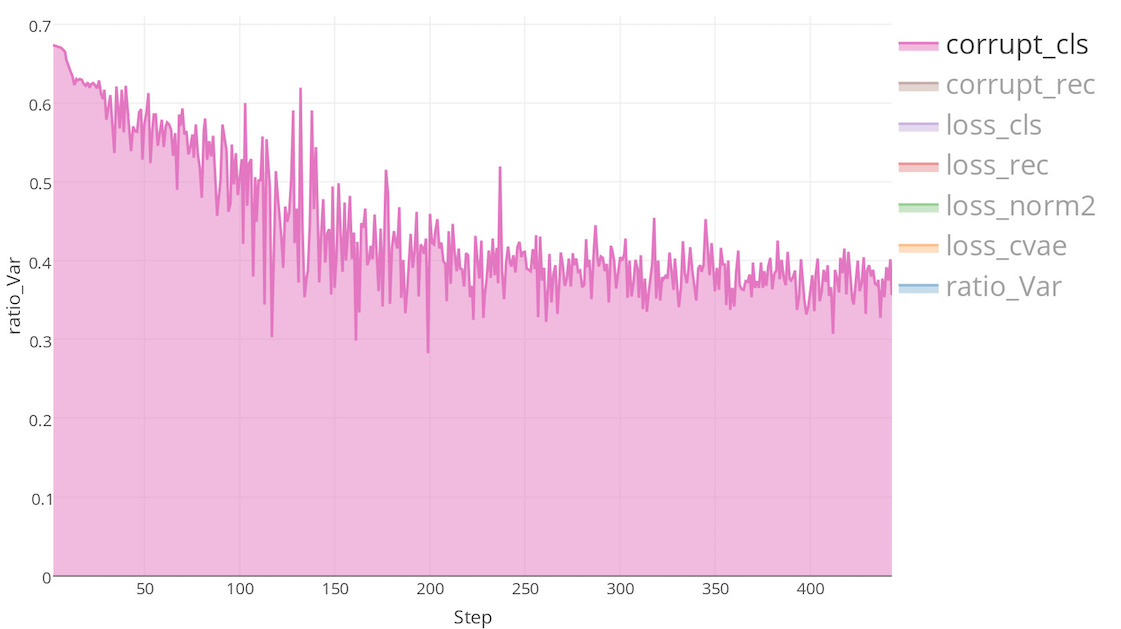}}
		\caption{classification noise ratio\label{fig:linechart:corrupcls}}
	\end{subfigure}
	\caption{Line charts of training parameters for experiments based on ShapeNet.\label{fig:linechart}}
\end{figure}

Here we explain how our generative model works in regarding to both corruption function (\ref{equ:noise_ratios}) and loss function (\ref{equ:loss_total}).
The conjugate network works as a whole in the training stage rather than gets trained separately.
Both KL divergence loss in Fig.~\ref{fig:linechart:losscvae} and the Euclidean loss in the reconstruction network in Fig.~\ref{fig:linechart:lossnorm2} are optimized in similar trends and converge in similar training stage.
There is a rapid drop of the KL divergence loss at the very beginning of training,
then the divergence keeps in a particular range for a long time while the Euclidean loss keeping going smaller. 
This phenomenon indicates that the encoder in our generative model helps to construct a latent space where the output of the encoder can provide discriminant information for the generator stably which makes it easier for training a reconstructor without over-fitting.

Noising ratio plays an important role for training concatenated network,
analysis on relationship of losses and noising ratio explains how adaptive noises work for tuning the whole parametric model.
Variation ratio between reconstructed images and semantic masks in Fig.~\ref{fig:linechart:ratio} gradually increases which means that discriminant information in reconstructed images increases during training because the information in expected semantic masks is almost the same.
Noising ratios used for processing realistic inputs of two sub-networks are shown in Fig.~\ref{fig:linechart:corruprec} for reconstruction network and Fig.~\ref{fig:linechart:corrupcls} for classification network. 
They are in negative and positive correlation with the variation ratio in Fig.~\ref{fig:linechart:ratio} accordingly which means that difficulty for training reconstruction network increases while training for classification network becoming easier during training.
Coupled noising ratios make the reconstruction sub-network utilize the back-propagated gradient from classification sub-network more efficiently.
At the same time, the classification sub-network utilizes all six input channels evenly without over-fitting on realistic RGB channels more easily than the reconstructed channels in the beginning. 
By deceasing information of input images for classification sub-network,
information in all six channels are always balanced during training.
Gradient from all six channels in classification network could be fully utilized in the beginning rather than only from three of them in synthetic images with background.
The intersection of three lines in Fig.~\ref{fig:linechart:losscls_corruprec_corrupcls} means that two noising ratios applied on reconstruction and classification networks equal to each other at this time.
Here it shows that loss of classification in Fig.~\ref{fig:linechart:losscls} is already decreased substantially and classification parametric model converges in a much smoother way in training afterwards without unstable loss turbulence meanwhile.
The variance ratio in Fig.~\ref{fig:linechart:ratio} keeps increasing monotonically with reconstruction ratio in Fig.~\ref{fig:linechart:corruprec} which means that realistic RBG images with background are not well-fitted on purpose before there are enough discriminant information given by the reconstructed channels.
Once the noising ratio of realistic images for classification network is smaller than noising ratio for reconstruction network, the reconstruction network tends to feed more information to classification network.
Although corruption for the input of reconstruction network in Fig.~\ref{fig:linechart:corruprec} keeps increasing, reconstruction loss in Fig.~\ref{fig:linechart:lossnorm2} is always decreasing which means that the generator keeps learning discriminant information even if the realistic input is more and more confusing.

\subsubsection{Advantages to Other Pixel-wise Reconstruction Methods}

\begin{figure}[ht]
	\captionsetup[subfigure]{justification=centering}
	\centering
	\begin{center}
		\begin{subfigure}[t]{0.95\linewidth}
			\centering
			\centerline{\includegraphics[width=1\linewidth]{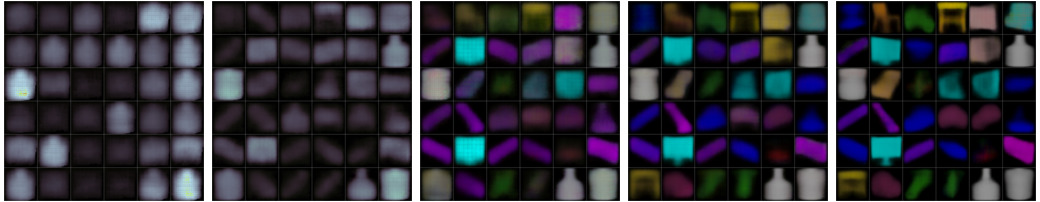}}
			\caption{Generative model with coordinate adaptive noises\label{fig:rec_corrup_noise}}
		\end{subfigure}
		\begin{subfigure}[t]{0.95\linewidth}
			\centering
			\centerline{\includegraphics[width=1\linewidth]{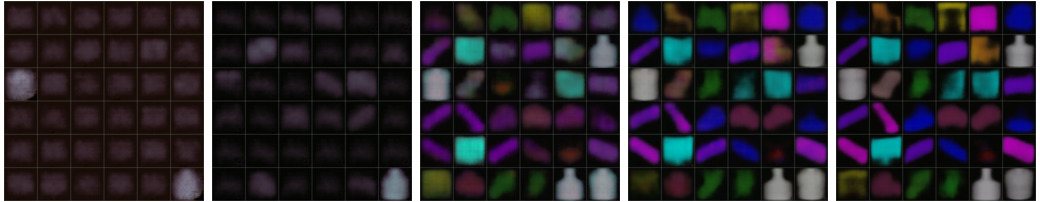}}
			\caption{Generative model with fixed noises\label{fig:rec_corrupALL}}
		\end{subfigure}
		\begin{subfigure}[t]{0.95\linewidth}
			\centering
			\centerline{\includegraphics[width=1\linewidth]{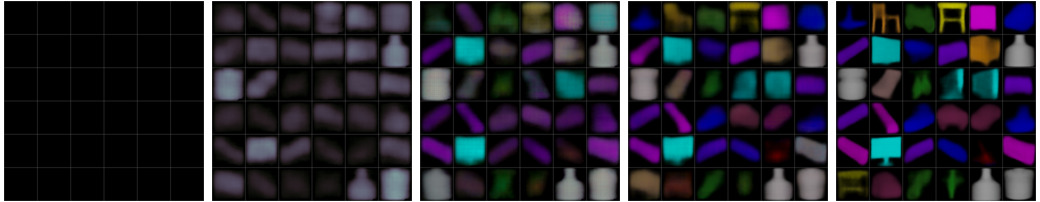}}
			\caption{Variational coder\label{fig:rec_cgm}}
		\end{subfigure}
		\begin{subfigure}[t]{0.95\linewidth}
			\centering
			\centerline{\includegraphics[width=1\linewidth]{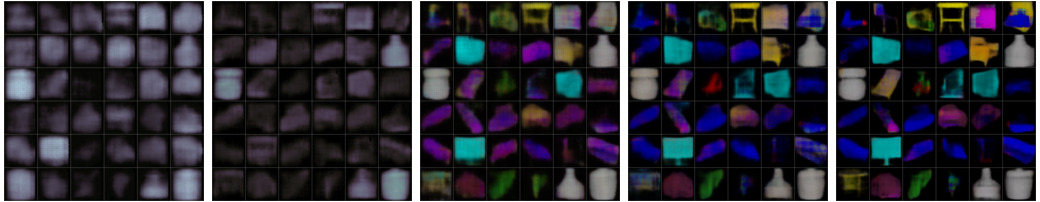}}
			\caption{FCN with adaptive noises\label{fig:rec_corrup_noise_fcn}}
		\end{subfigure}
		\begin{subfigure}[t]{0.95\linewidth}
			\centering
			\centerline{\includegraphics[width=1\linewidth]{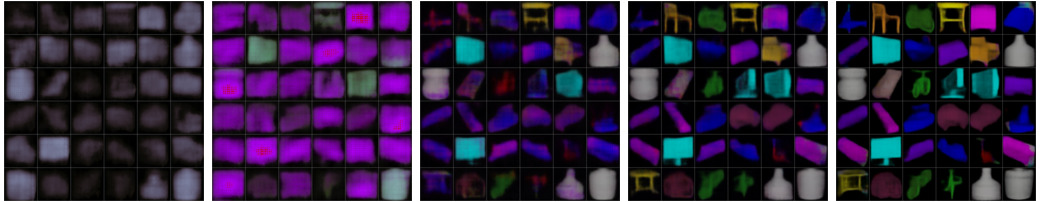}}
			\caption{FCN\label{fig:rec_fcn}}
		\end{subfigure}
		\caption{Visualization of reconstructed samples according to inputs and randomly selected latent variables.\label{fig:process_compare}}
	\end{center}
\end{figure}

As reconstruction sub-network of our conjugate architecture depends on variational inference,
the final reconstruction result is similar to the one learned from CVAE~\cite{NIPS2015_5775,Kingma2014Semi}.
The advantage of our architecture compared to simple variational coder (VC) is shown in the learning process in Fig.~\ref{fig:rec_corrup_noise} and Fig.~\ref{fig:rec_cgm} where reconstruction sub-network learns faster than VC due to the classification sub-network.
Our conjugate generative model also performs better than other popular pixel-wise regression models such as FCN~\cite{Long2015Fully,Wang2015Visual} and DeconvNet~\cite{Noh2015Learning} if the final one-hot coding layer is replaced by pixel-wise regression layer with 3 channels.
As shown in Fig.~\ref{fig:process_compare}, our generative model with metric learning (GM-ML) has advantages both on the speed of learning and the precision of reconstruction accuracy. 
Results of methods like DeconvNet shows that directly doing a pixel-wise coding from realistic images to semantic masks does not work well because FCN and DeconvNet treat semantic segmentation tasks without probabilistic variables in latent space,
direct learning from input makes it less robust to noises for FCN\@.
Firstly, FCN reconstruction result shown in the first two images of Fig.~\ref{fig:rec_fcn} shows over fitting problem for synthetic data where the local contours are learned at first without forming global shapes.
Secondly, final result of FCN for the last image in Fig.~\ref{fig:rec_fcn} can not render monotone RGB colors according to unique categories where a single object has more than one colors on the surface.
Our GM-ML makes it possible to train a pixel-wise coding strategy based on paired input and expected output with hidden meaning of category by forming clusters in the latent space.
Our method shown in Fig.~\ref{fig:rec_corrup_noise} indicates that it fits for the learning target shown in Fig.~\ref{fig:recon_train_mask} well.
If noises are fixed rather than adaptively changed according to output of reconstruction channels like result in the first two images of Fig.~\ref{fig:rec_corrupALL}, the learning progress is much slower.

\subsubsection{Feature Distributions, Reconstruction and Classification}

\begin{figure}[ht]
	\captionsetup[subfigure]{justification=centering}
	\centering
	\begin{subfigure}[t]{0.19\linewidth}
		\centering
		\centerline{\includegraphics[width=1\linewidth]{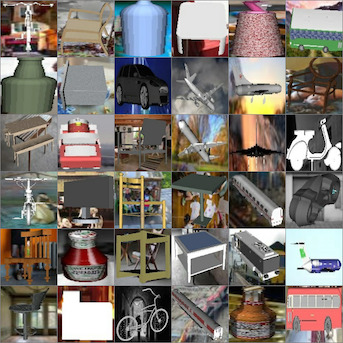}}
		\caption{Input\label{fig:recon_shapenet_input}}
	\end{subfigure}
	\begin{subfigure}[t]{0.19\linewidth}
		\centering
		\centerline{\includegraphics[width=1\linewidth]{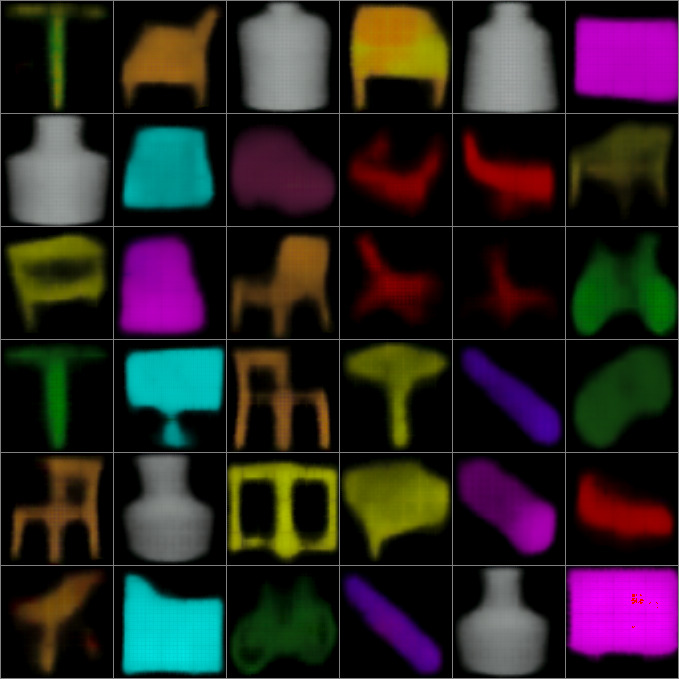}}
		\caption{VC\label{fig:recon_shapenet_cvae}}
	\end{subfigure}
	\begin{subfigure}[t]{0.19\linewidth}
		\centering
		\centerline{\includegraphics[width=1\linewidth]{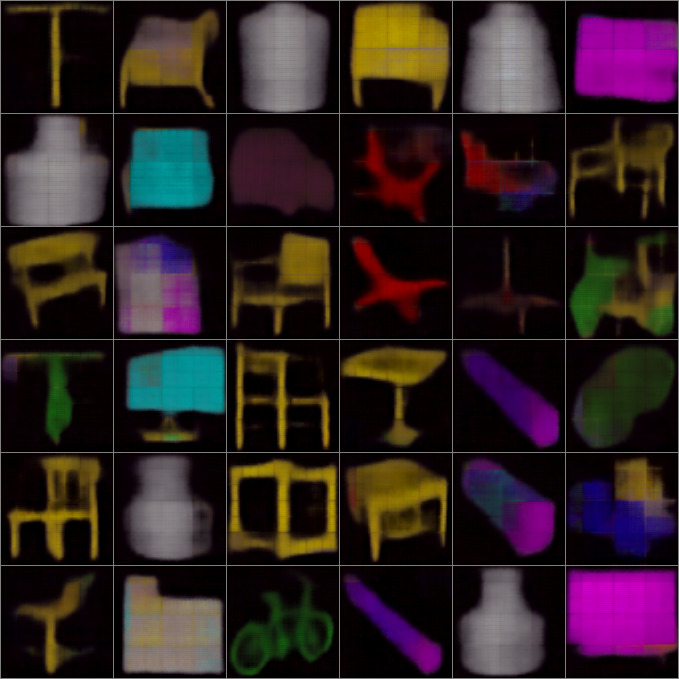}}
		\caption{FCN\label{fig:recon_shapenet_fcn}}
	\end{subfigure}
	\begin{subfigure}[t]{0.19\linewidth}
		\centering
		\centerline{\includegraphics[width=1\linewidth]{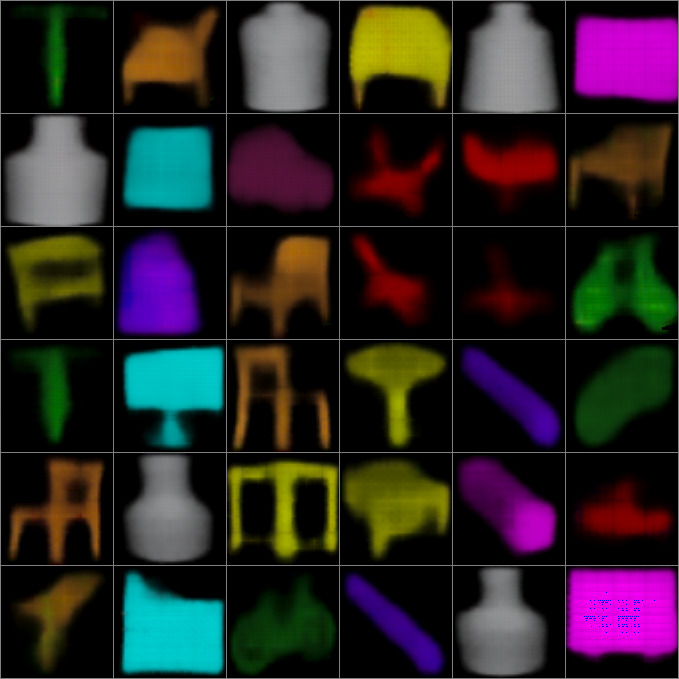}}
		\caption{GM-ML\label{fig:recon_shapenet_cgm}}
	\end{subfigure}
	\begin{subfigure}[t]{0.19\linewidth}
		\centering
		\centerline{\includegraphics[width=1\linewidth]{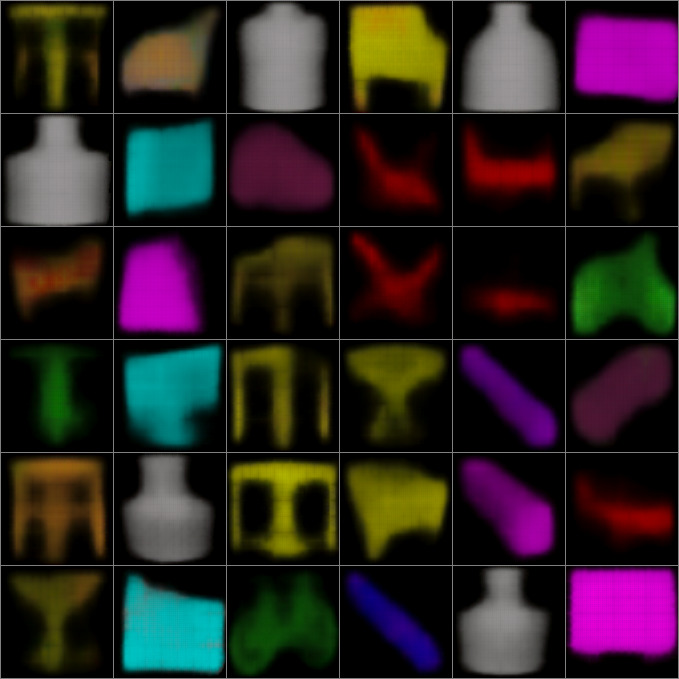}}
		\caption{GM-CML\label{fig:recon_shapenet_adcgm}}
	\end{subfigure}
	\caption{Reconstruction samples of synthetic images rendered from 3D models of ShapeNet. VC is variational coder with flexible output. GM-CML (Generative Model with Coordinate Metric Learning) indicates that the conjugate generative model is trained with adaptive noises and GM-CML represent the same architecture without adaptive noises.\label{fig:recon_shapenet}}
\end{figure}

\begin{figure}[ht]
	\captionsetup[subfigure]{justification=centering}
	\centering
	\begin{subfigure}[t]{0.46\linewidth}
		\centering
		\centerline{\includegraphics[width=1\linewidth]{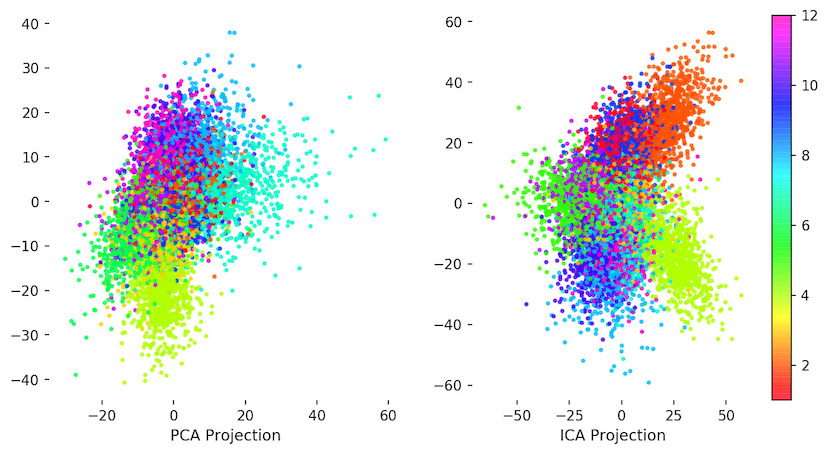}}
		\caption{VC\label{fig:dist_shapenet_cvae}}
	\end{subfigure}
	\begin{subfigure}[t]{0.46\linewidth}
		\centering
		\centerline{\includegraphics[width=1\linewidth]{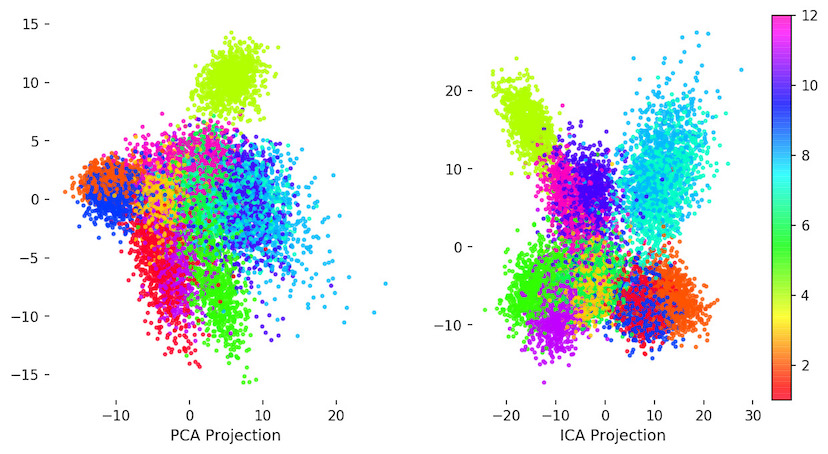}}
		\caption{FCN\label{fig:dist_shapenet_fcn}}
	\end{subfigure}
	\begin{subfigure}[t]{0.46\linewidth}
		\centering
		\centerline{\includegraphics[width=1\linewidth]{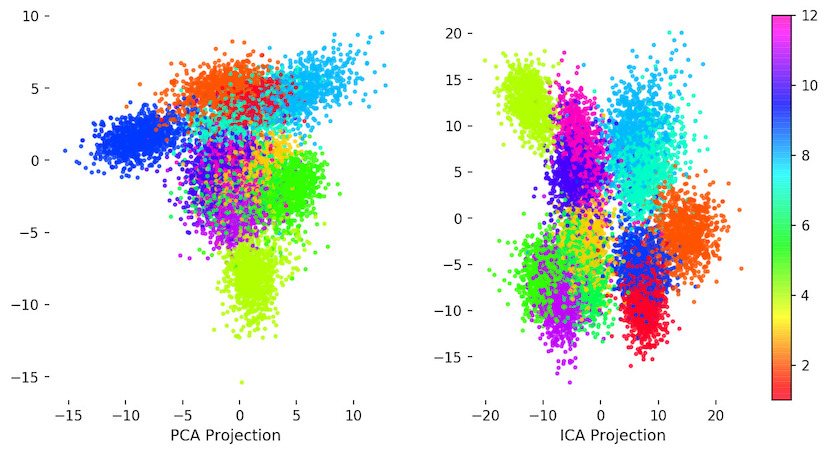}}
		\caption{GM-ML\label{fig:dist_shapenet_cgm}}
	\end{subfigure}
	\begin{subfigure}[t]{0.46\linewidth}
		\centering
		\centerline{\includegraphics[width=1\linewidth]{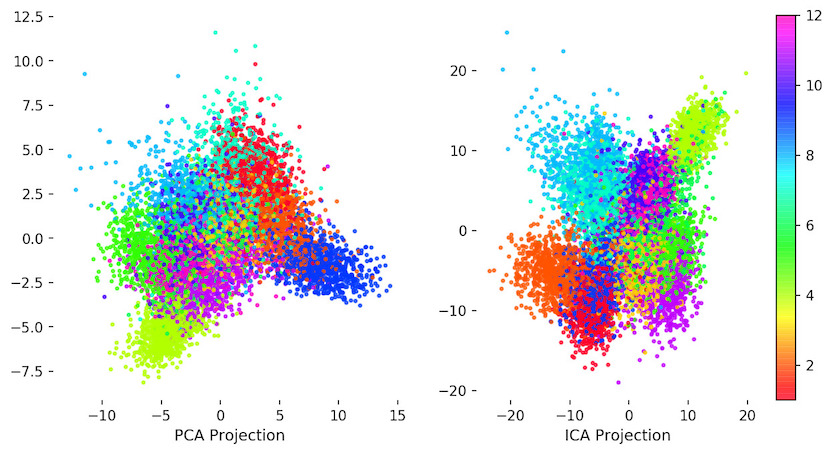}}
		\caption{GM-CML\label{fig:dist_shapenet_adcgm}}
	\end{subfigure}
	\caption{Two-dimensional projection using PCA and ICA.\label{fig:dist_shapenet}}
\end{figure}

We visualize the distribution of descriptors by projecting the output of classification sub-network to two dimensions using principal component analysis (PCA)~\cite{jolliffe2014principal} and independent component analysis (ICA).
Distribution shown in Fig.~\ref{fig:dist_shapenet_adcgm} indicates that our generative model trained with adaptive noises has a desired distribution of descriptors for classification where descriptors in different class are separated evenly.
Projected features of VC in Fig.~\ref{fig:dist_shapenet_cvae} can not identify samples according to their categories as good as ours shown in Fig.~\ref{fig:dist_shapenet_cgm} where samples from the same category are better clustered.
This comparison experiment shows that directly concatenating a classification network to generative model without back-propagating gradients from classification sub-network makes it hard to fully utilize the reconstructed channels due to the absence of categories information.
Here the reconstruction layer could be regarded as both a reconstruction target and a bridge providing additional supervision information.
The variational variables also matter much in the whole pipeline, reconstructed foreground objects using FCN~\cite{Long2015Fully} concatenated with a classification network shown in Fig.~\ref{fig:recon_shapenet_fcn} have more blended colors than ours in Fig.~\ref{fig:recon_shapenet_cgm} where foreground objects are rendered with monotone colors.
This means that the pixel-wise reconstructor can not render foreground objects stably according to categories without using generative model even when it can also reconstruct contours well.

Table.~\ref{tab:shapenet} shows classification results for testing on the same type of synthetic realistic images rendered from ShapeNet~\cite{shapenet2015} with different backgrounds and object poses compared to training data.
Our conjugate deep architecture is also more capable for classification tasks than other popular deep architectures due to the help of the foreground reconstruction sub-network.
Table.~\ref{tab:shapenet} shows that our generative model (GM-ML\_{1}) of which output of the reconstruction sub-network is concatenated as a part of the input of ZigzagNet for classification achieves the accuracy of 97.9\% which is 7.8\% higher than the one trained from ZigzagNet.
Such improvement shows that the additional 3 channels of semantic foreground object extraction helps a lot for improving the classification network when data migration problem is not severe.
Comparison experiments on different reconstruction sub-networks show that generative model retrieves the hidden category information with 5.1\% improvement on classification accuracy compared to FCN\@.

\begin{table}[ht]
	\renewcommand{\arraystretch}{1.3}
	\caption{Classification on rendered samples from ShapeNet database. Both training and testing data are synthetic images rendered from models of ShapeNet database in 12 categories. GM-CML$_1$ means that our conjugate generative model is trained with adaptive noises and the output of reconstruction network is concatenated with the input of ZigzagNet. GM-ML$_2$ means that the classification network is AlexNet.\label{tab:shapenet}}
	\centering
	\begin{tabular}{|c||c||c|}
		\hline
		\diagbox{\bfseries Method}{\bfseries Result} & \bfseries Accuracy & \bfseries Model size \\
		\hline\hline
		GM-CML$_1$                                   & 96.3\%             & 20.2 MB              \\
		GM-ML$_1$                                    & \textbf{97.9}\%             & 20.2 MB              \\
		GM-ML$_2$                                    & 97.1\%             & 20.2 MB              \\
		FCN~\cite{Long2015Fully}                     & 91.2\%             & 20.2 MB              \\
		DeconvNet~\cite{Noh2015Learning}             & 92.7\%             & 20.2 MB              \\
		AlexNet~\cite{Krizhevsky2012}                & 86.2\%             & 240 MB               \\
		SqueezeNet~\cite{Forrest16CoRR}              & 84.7\%             & 4.8 MB               \\
		ZigzagNet~\cite{wang_accv2016}               & 90.1\%             & 6.2 MB               \\
		\hline
	\end{tabular}
\end{table}

\subsection{Experiments on ImageNet Database}

\begin{figure}[ht]
	\captionsetup[subfigure]{justification=centering}
	\centering
	\begin{subfigure}[t]{0.19\linewidth}
		\centering
		\centerline{\includegraphics[width=1\linewidth]{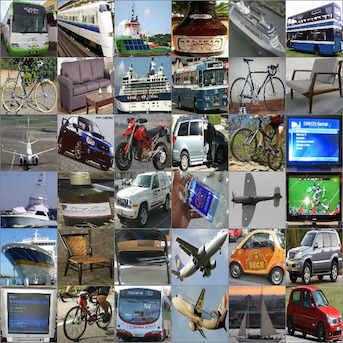}}
		\caption{Input\label{fig:recon_imagenet_input}}
	\end{subfigure}
	\begin{subfigure}[t]{0.19\linewidth}
		\centering
		\centerline{\includegraphics[width=1\linewidth]{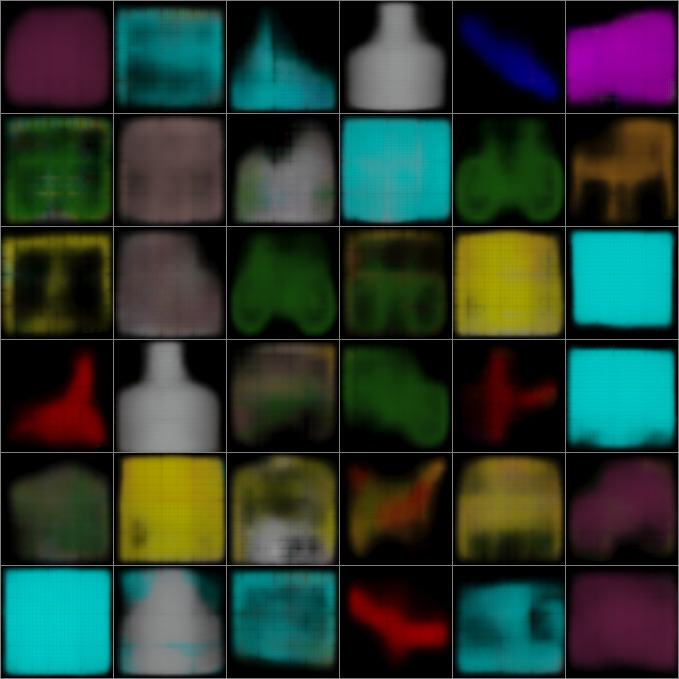}}
		\caption{VC\label{fig:recon_imagenet_cvae}}
	\end{subfigure}
	\begin{subfigure}[t]{0.19\linewidth}
		\centering
		\centerline{\includegraphics[width=1\linewidth]{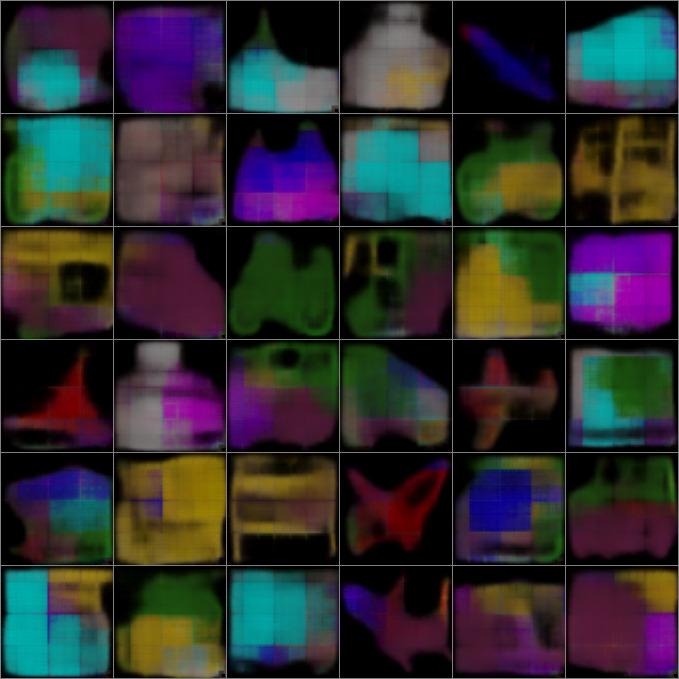}}
		\caption{FCN\label{fig:recon_imagenet_fcn}}
	\end{subfigure}
	\begin{subfigure}[t]{0.19\linewidth}
		\centering
		\centerline{\includegraphics[width=1\linewidth]{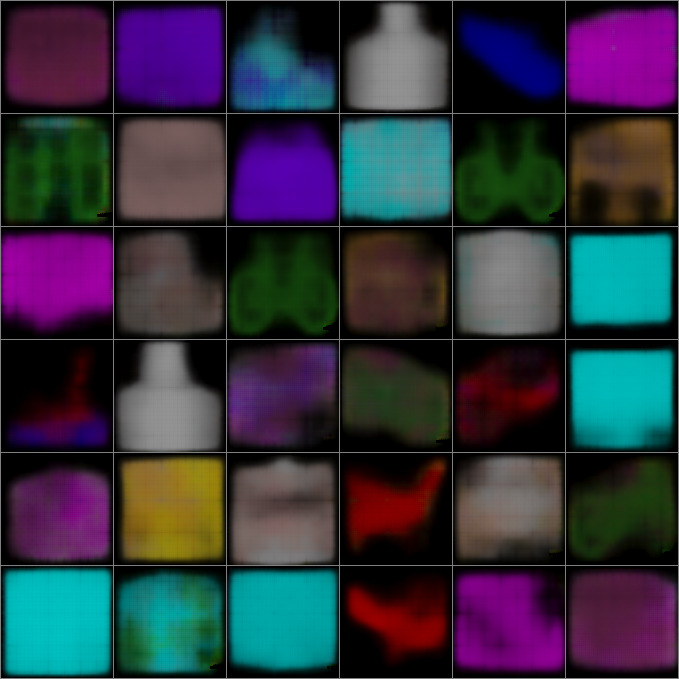}}
		\caption{GM-ML\label{fig:recon_imagenet_cgm}}
	\end{subfigure}
	\begin{subfigure}[t]{0.19\linewidth}
		\centering
		\centerline{\includegraphics[width=1\linewidth]{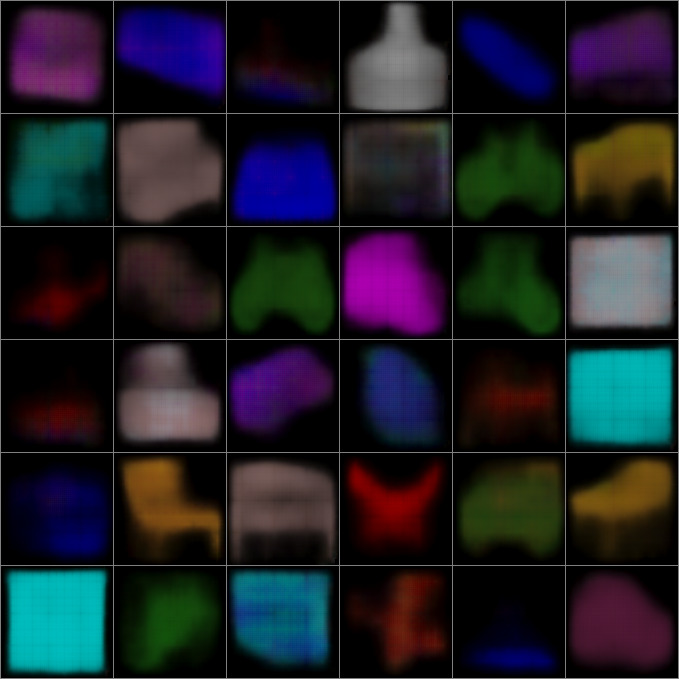}}
		\caption{GM-CML\label{fig:recon_imagenet_adcgm}}
	\end{subfigure}
	\caption{Reconstruction samples of real images from ImageNet.\label{fig:recon_imagenet}}
\end{figure}

\begin{figure}[ht]
	\begin{subfigure}[t]{0.46\linewidth}
		\centering
		\centerline{\includegraphics[width=1\linewidth]{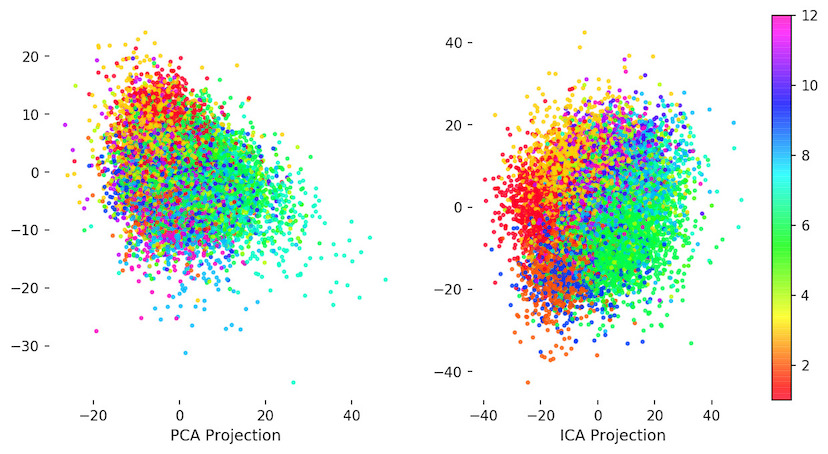}}
		\caption{VC\label{fig:dist_imagenet_cvae}}
	\end{subfigure}
	\begin{subfigure}[t]{0.46\linewidth}
		\centering
		\centerline{\includegraphics[width=1\linewidth]{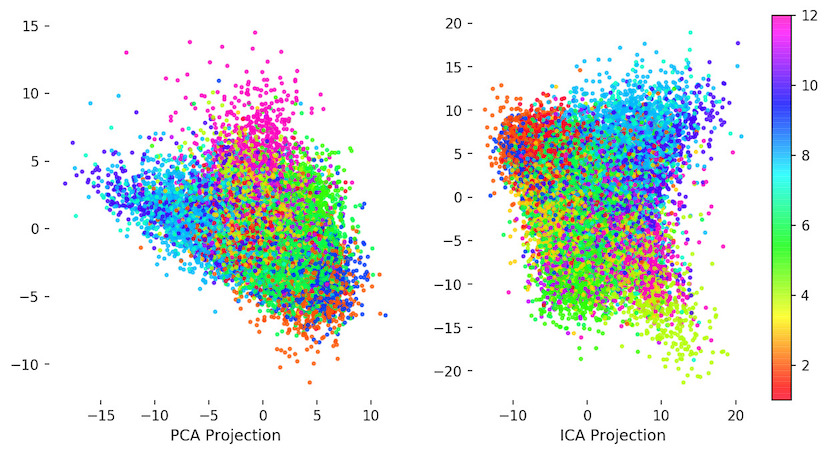}}
		\caption{FCN\label{fig:dist_imagenet_fcn}}
	\end{subfigure}

	\begin{subfigure}[t]{0.46\linewidth}
		\centering
		\centerline{\includegraphics[width=1\linewidth]{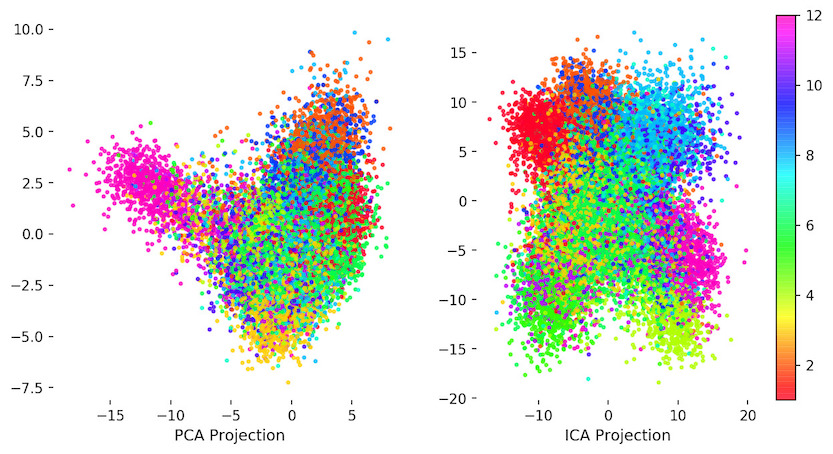}}
		\caption{GM-ML\label{fig:dist_imagenet_cgm}}
	\end{subfigure}
	\begin{subfigure}[t]{0.46\linewidth}
		\centering
		\centerline{\includegraphics[width=1\linewidth]{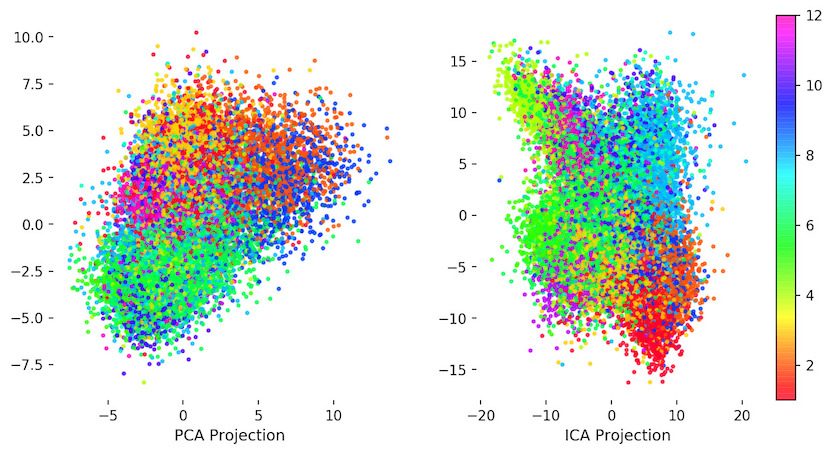}}
		\caption{GM-CML\label{fig:dist_imagenet_adcgm}}
	\end{subfigure}
	\caption{Two-dimensional projection using PCA and ICA.\label{fig:dist_imagenet}}
\end{figure}

Here real photos of PASCAL and ImageNet are part of the original database which are provided by PASCAL 3D+~\cite{xiang_wacv14} database with 12 categories altogether.
Those experiments are designed for proving that our method works well for testing on real photos with the parametric model trained by realistic data rendered by 3D models.
We do classification using outputs from softmax classifier which is the last layer in classification network to evaluate the performance.
We render training data from models of PASCAL 3D+ with rendering method of SR~\cite{wang_icip2016} for basic experiments and models of ShapeNet database with rendering method in ZigzagNet.
Some experiment results are collected from original papers if our testing results are similar to previously declared ones.
Experiment in Table.~\ref{tab:imagenet_tip} on trained network for regular photos from ImageNet attached in PASCAL 3D+~\cite{Olga2014} shows that conjugating generative model with the help of adaptive noises could extract semantic foreground object information from background better compared to other popular and recent networks,
achieving the state of the art accuracy of 50.5\% for real photos from ImageNet.
Comparison experiments on other reconstruction methods such as FCN~\cite{Long2015Fully} and DCN~\cite{Noh2015Learning} show that latent variables in our generative model are helpful for extracting categorical informations in the foreground semantic masks which contribute to about 10\% accuracy rising.
We use the same macro reconstruction architecture for our conjugate generative model, FCN and DeconvNet, the main difference in other reconstruction sub-network for comparison is that they do not have the latent variables.
Visualization of the reconstructed channels in Fig.~\ref{fig:recon_imagenet_adcgm} shows that our method has advantage in revealing details of objects compared to object segmentation results without texture information and can also suppressing close-up background compared to depth images.
Here object reconstruction is applicable in extracting objects in real images by distinguishing object from background while suppressing highlights regions in background.
Although foreground objects reconstructed in form of 1-channel depth images in SR~\cite{wang_icip2016} could give a help on doing classification on real photos by means of concatenating reconstructed channel to RGB channels, semantic colors are missing.

Our compact generative model based on ZigzagNet could be stored on disk occupying about 20.2 MB including both reconstruction network and classification network and achieves highest accuracy compared to both SqueezeNet~\cite{Forrest16CoRR} and AlexNet.

\begin{table}[ht]
	\renewcommand{\arraystretch}{1.3}
	\caption{Classification on ImageNet samples attached in PASCAL 3D+ database. Different types of data used for training are real photos (1st column) and synthetic images which are rendered from PASCAL 3D+ (2nd column) and ShapeNet (3rd column) models. GM-CML$_1$ means that our conjugate generative model is trained with adaptive noises and the output of reconstruction network is concatenated with the input of ZigzagNet. GM-ML$_2$ means that the classification network is AlexNet.\label{tab:imagenet_tip}}
	\centering
	\begin{tabular}{|c||c||c||c|}
		\hline
		\diagbox{\bfseries Method}{\bfseries Data} & \bfseries Real photo & \bfseries PASCAL 3D & \bfseries ShapeNet \\
		\hline
		\hline
		GM-CML$_1$                                 & \textbf{\textit{63.1}}\% & \textbf{27.8}\%             & \textbf{50.5}\%             \\
		GM-ML$_1$                                  & \textit{58.6}\% & 26.1\%             & 49.2\%             \\
		GM-ML$_2$                                  & \textit{60.9}\% & 25.1\%             & 48.2\%             \\
		FCN~\cite{Long2015Fully}                   & \textemdash & 23.9\%             & 38.4\%             \\
		DeconvNet~\cite{Noh2015Learning}           & \textemdash & 26.2\%             & 40.2\%             \\
		SR~\cite{wang_icip2016}                    & \textemdash & 25.1\%             & 40.1\%             \\
		AlexNet~\cite{Krizhevsky2012}              & 55.8\%      & 24\%               & 39.6\%             \\
		SqueezeNet~\cite{Forrest16CoRR}            & 45.5\%      & 21.4\%             & 35\%               \\
		ZigzagNet~\cite{wang_accv2016}             & 55.9\%      & 25.1\%             & 42.7\%             \\
		Inception V1~\cite{SzegedyLJSRAEVR14}      & 59\%        & 21.9\%             & 29.9\%             \\
		Inception V4~\cite{SzegedyIV16}            & \textemdash & 23.7\%             & 37.2\%             \\
		ResidualNet~\cite{he2015deep}              & \textemdash & 19.1\%             & 28.5\%             \\
		LDO~\cite{wohlhart15}                      & \textemdash & 14.8\%             & 25.5\%             \\
		\hline
	\end{tabular}
\end{table}

\subsection{Experiments on Pascal 12 Database}

We further test our generative model on a difficult test set provided by PASCAL 3D+ database. Test images selected from PASCAL 2012~\cite{everingham2015the} are much more complicated than ones from ImageNet where foreground objects are not centered and shot as a whole part.
This means that parts of object without complete contour are captured in sample photos. Basic result is carried out on texture-less models provided in PASCAL 3D+ database which shows that a severe data migration problem makes it too hard to get a good performance no matter which network we choose where all accuracies are not higher than 20\% in Table.~\ref{tab:pascal_tip}.
It's so necessary to use synthetic images rendered from textured model provided in ShapeNet database instead of ones from texture-less models in PASCAL 3D+ in such condition with incomplete foreground objects waiting to be recognized.
Parametric network trained from ShapeNet models has 10\% improvement on accuracy with the help of reconstructed foreground objects.
As objects are not centered in test photos, three reconstructed channels help to reconstruct foreground objects and improve recognition accuracy to 29.7\% which is already much higher than that from deep models like Inception V4~\cite{SzegedyIV16} which is only 20.1\%.
This phenomenon means that deep residual architectures like ResidualNet~\cite{he2015deep} and Inception V4~\cite{SzegedyIV16} with strong fitting capability still have their weakness when training on synthetic data and testing on real photos with data migration problem without the help of pose information.
The reason why direct classification networks like AlexNet without reconstruction sub-network fail to achieve acceptable accuracy as our model in complex condition is that our coordinate generative is assisted with metric learning targeting on utilizing common knowledge rather than over-fitting all information.
This means that our generative model has its own limitation on fitting with input data,
but the whole parametric model is tuned on fundamental knowledge like relative pose information which could be easily collected from synthetic data.
As show in Table.~\ref{tab:pascal_tip}, the most amazing point is that the accuracy of our pipeline is nearly twice as high as ones trained without generative model which are 15.2\% for FCN~\cite{Long2015Fully} concatenated with a classification network and 15.1\% for DeconvNet\cite{Noh2015Learning}.
Obvious advantage in classification against traditional pixel-wise reconstruction methods could be explained by reconstruction results shown in Fig.~\ref{fig:recon_imagenet} where our method avoids problem of tearing foreground objects up into several different objects in form of rendering it with different colors in Fig.~\ref{fig:recon_imagenet_fcn}.
Close accuracy between our generative model and ZigzagNet indicates that pose information helps to enhance the ability of perceiving ground truth relative information which avoid over-fitting for uncommon or random category information in test samples.
Here we can draw conclusion that our generative model with coordinate metric learning could try best to reconstruct foreground objects in a global view rather than a pixel-wise view like FCN where the surface colors are always pure,
so over-fitting problem is avoided by not reconstructing clear foreground object object channels.

\begin{table}[ht]
	\renewcommand{\arraystretch}{1.3}
	\caption{Classification on Pascal 2012 samples attached in PASCAL 3D+ database. Different types of data used for training are real photos (1st column) and synthetic images which are rendered from PASCAL 3D+ (2nd column) and ShapeNet (3rd column) models. GM-CML$_1$ means that our conjugate generative model is trained with adaptive noises and the output of reconstruction network is concatenated with the input of ZigzagNet. GM-ML$_2$ means that the classification network is AlexNet.\label{tab:pascal_tip}}
	\centering
	\begin{tabular}{|c||c||c||c|}
		\hline
		\diagbox{\bfseries Method}{\bfseries Data} & \bfseries Real photo & \bfseries PASCAL 3D & \bfseries ShapeNet \\
		\hline\hline
		GM-CML$_1$                                 & \textbf{\textit{46.7}} & \textbf{18.1}\%             & \textbf{29.7}\%             \\
		GM-ML$_1$                                  & \textit{42.1} & 18\%               & 29.4\%             \\
		GM-ML$_2$                                  & \textit{44.5} & 17.2\%             & 28\%               \\
		FCN~\cite{Long2015Fully}                   & \textemdash & 13.4\%             & 15.2\%             \\
		DeconvNet~\cite{Noh2015Learning}           & \textemdash & 12.1\%             & 15.1\%             \\
		SR~\cite{wang_icip2016}                    & \textemdash & 13.3\%             & 16.9\%             \\
		AlexNet~\cite{Krizhevsky2012}              & 39.2\%      & 15.2\%             & 20.7\%             \\
		SqueezeNet~\cite{Forrest16CoRR}            & 32.9\%      & 14.8\%             & 19.6\%             \\
		ZigzagNet~\cite{wang_accv2016}             & \textbf{42.9}\%      & 16.7\%             & 27.9\%             \\
		Inception V1~\cite{SzegedyLJSRAEVR14}      & 42.3\%      & 14.7\%             & 17.3\%             \\
		Inception V4~\cite{SzegedyIV16}            & \textemdash & 16.7\%             & 20.1\%             \\
		ResidualNet~\cite{he2015deep}              & \textemdash & 12.3\%             & 16.2\%             \\
		LDO~\cite{wohlhart15}                      & \textemdash & 11.2\%             & 18.7\%             \\
		\hline
	\end{tabular}
\end{table}

As we have achieve the state of the art performance based for training with synthetic data alone and testing on real photos. We further carry out experiments for our methods in Table~\ref{tab:imagenet_tip} and Table~\ref{tab:pascal_tip} to prove that our pre-trained model can also improve the performance of fine-tuning with real data.
Accuracies shown in \textit{italic} font in Table~\ref{tab:imagenet_tip} and Table~\ref{tab:pascal_tip} represent that our synthetic data are used for pre-training and real data are used for fine-tuning.
Most of those results are higher than results trained directly by real photos which shows that our parametric model pre-trained by synthetic data could improve the final performance if real images are used as training data for fine-tuning afterwards.

\section{Conclusion}
We designed a unified deep generative model with two concatenated sub-networks to do foreground object reconstruction and categorical classification with metric learning at the same time.
We try to solve the ultimate target of training deep parametric model barely on synthetic images rendered from 3D models and testing on real images.
A coordinate training strategy based on variance ratio makes our conjugate network work effectively where two concatenated sub-networks converge quicker and benefit more from each others.
Experiments on real photos from ImageNet database attached in ShapeNet database shows that our generative model with coordinate metric learning achieves the state of the art classification accuracy of 50.5\% trained on rendered data.
The generative model also shows more flexible reconstruction ability compared to VAE while removing additional supervision dependency of CVAE\@.
From perspective of generative model, it gains faster converging speed compared to variational coder.

\section*{Acknowledgment}
This work was partially sponsored by supported by the NSFC (National Natural Science Foundation of China) under Grant No. 61375031, No. 61573068, No. 61471048, and No.61273217, the Fundamental Research Funds for the Central Universities under Grant No. 2014ZD03\textemdash 01, This work was also supported by Beijing Nova Program, CCF\textemdash Tencent Open Research Fund, and the Program for New Century Excellent Talents in University.


\ifCLASSOPTIONcaptionsoff
\fi

\bibliographystyle{IEEEtran}
\bibliography{wang_gmcml_2017}

%

\begin{IEEEbiography}[{\includegraphics[width=1in,height=1.25in,clip,keepaspectratio]{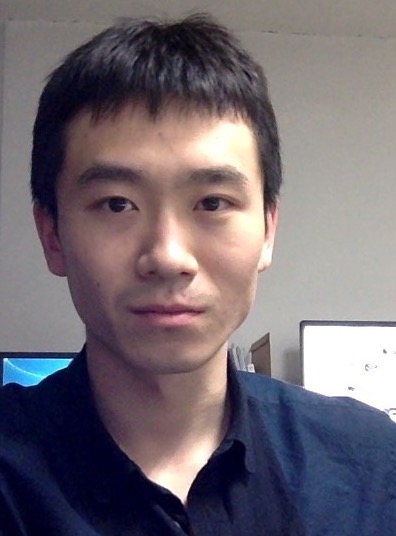}}]{Yida Wang}
	is Ph.D candidate in Technical University of Munich, Munich, Germany. He received B.E. and M.E. degrees from Beijing University of Posts and Telecommunications, Beijing, China in 2014 and 2017, respectively. His research interests include pattern recognition and computer vision. He was invited by Microsoft Research Seattle for Microsoft Faculty Summit in 2016 for project of ``CNTK on Mac: 2D Object Restoration and Recognition Based on 3D Model'' which was awarded as the second prize for Microsoft Open Source Challenge 2016. He is currently sponsored by Bleenco Research Fellowship in 2018 and was also sponsored by Google Summer of Code project twice for deep learning projects together with OpenCV organization. He was named of Excellent Graduate Student of Beijing City twice in 2014 and 2017 and been awarded for National Scholarship for Graduate Students in 2016.
\end{IEEEbiography}


\begin{IEEEbiography}[{\includegraphics[width=1in,height=1.25in,clip,keepaspectratio]{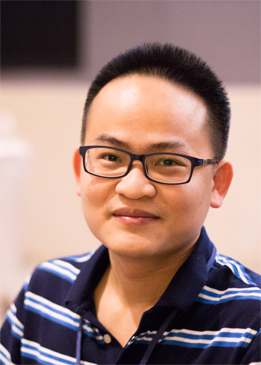}}]{Weihong Deng}
	received the B.E. degree in information engineering and the Ph.D. degree in signal and information processing from the Beijing University of Posts and Telecommunications (BUPT), Beijing, China, in 2004 and 2009, respectively. From Oct. 2007 to Dec. 2008, he was a postgraduate exchange student in the School of Information Technologies, University of Sydney, Australia. He is currently an professor in School of Information and Telecommunications Engineering, BUPT\@. His research interests include statistical pattern recognition and computer vision, with a particular emphasis in face recognition. He has published over 100 technical papers in international journals and conferences, such as IEEE TPAMI and CVPR\@. He serves as associate editor for IEEE Access, and guest editor for Image and Vision Computing Journal and the reviewer for dozens of international journals, such as IEEE TPAMI / TIP / TIFS / TNNLS / TMM / TSMC, IJCV, PR / PRL\@. His Dissertation titled “Highly accurate face recognition algorithms” was awarded the Outstanding Doctoral Dissertation Award by Beijing Municipal Commission of Education in 2011. He has been supported by the program for New Century Excellent Talents by the Ministry of Education of China in 2013 and Beijing Nova Program in 2016.
\end{IEEEbiography}




\end{document}